%% file: snorkel_vldb17.tex
\documentclass{vldb}
\usepackage{snorkel_vldb17}
\usepackage{fancybox, graphicx}
\usepackage{balance}  
\usepackage{mathtools}

\vldbTitle{Snorkel: Rapid Training Data Creation with Weak Supervision}
\vldbAuthors{A. Ratner, S. H. Bach, H. Ehrenberg, J. Fries, S. Wu, C. R\'e}
\vldbDOI{10.14778/3157794.3157797}

\newcommand{\userstudyspeedup}[0]{2.8} 
\newcommand{\userstudyimprovement}[0]{45.5} 
\newcommand{\optimizerGMadvacc}[0]{2.16} 
\newcommand{\optimizerGMmaxspeedup}[0]{1.8} 
\newcommand{\optimizerSLmaxtimesaved}[0]{61} 
\newcommand{\optimizerSLmaxtimesavedmin}[0]{34} 
\newcommand{\avgimprovementoverheuristic}[0]{132} 
\newcommand{\avgimprovementoverheuristicRW}[0]{110} 
\newcommand{\avgimprovementoverheuristicBM}[0]{153} 
\newcommand{\avgrelimprovementoverMV}[0]{5.81} 
\newcommand{\avgHLgap}[0]{3.60} 
\newcommand{\avgHLgapRE}[0]{2.11} 
\newcommand{\avgHLgapCM}[0]{5.08} 
\newcommand{\avgrecallimprovementDM}[0]{43.15} 

\begin{document}


\title{Snorkel: Rapid Training Data Creation \\ with Weak Supervision}


%
%
%
%

\numberofauthors{1} 

\author{
%
%
\alignauthor
Alexander Ratner~~~~~Stephen H. Bach~~~~~Henry Ehrenberg\\
\vskip 0.2em
Jason Fries~~~~~Sen Wu~~~~~Christopher R\'e\\
\vskip 0.2em
       \affaddr{Stanford University}\\
       \affaddr{Stanford, CA, USA}\\
       \email{\{ajratner, bach, henryre, jfries, senwu, chrismre\}@cs.stanford.edu}
}

\maketitle

\input{abstract}

\input{intro}

\input{system}

\input{modeling}

\input{experiments}

\input{related}

\input{conclusion}

\balance
\vskip .5em \noindent
{\small{\bf Acknowledgments.}
Alison Callahan and Nigam Shah of Stanford, and Nicholas Giori of the US Dept. of Veterans Affairs developed the electronic health records application.
Emily Mallory, Ambika Acharya, and Russ Altman of Stanford, and Roselie Bright and Elaine Johanson of the US Food and Drug Administration developed the scientific articles application.
Joy Ku of the Mobilize Center organized the user study.
Nishith Khandwala developed the radiograph application.
We thank the contributors to Snorkel including Bryan He, Theodoros Rekatsinas, and Braden Hancock.
We gratefully acknowledge the support of DARPA under No. N66001-15-C-4043 (SIMPLEX), No. FA8750-17-2-0095 (D3M), No. FA8750-12-2-0335, and No. FA8750-13-2-0039,
DOE 108845,
NIH U54EB020405,
ONR under No. N000141210041 and No. N000141310129,
the Moore Foundation,
the Okawa Research Grant,
American Family Insurance,
Accenture,
Toshiba,
the Stanford Interdisciplinary Graduate and Bio-X fellowships,
and members of the Stanford DAWN project: Intel, Microsoft, Teradata, and VMware.
The U.S. Government is authorized to reproduce and distribute reprints for Governmental purposes notwithstanding any copyright notation thereon. 
Any opinions, findings, and conclusions or recommendations expressed in this material are those of the authors and do not necessarily reflect the views, policies, or endorsements, either expressed or implied, of DARPA, DOE, NIH, ONR, or the U.S. Government.

\bibliographystyle{abbrv}
\small{\bibliography{snorkel_vldb17}}

\input{appendix}

\end{document}

%% file: abstract.tex

\begin{abstract}
Labeling training data is increasingly the largest bottleneck in deploying machine learning systems.
We present Snorkel, a first-of-its-kind system that enables users to train state-of-the-art models without hand labeling any training data.
Instead, users write labeling functions that express arbitrary heuristics, which can have unknown accuracies and correlations.
Snorkel denoises their outputs without access to ground truth by incorporating the first end-to-end implementation of our recently proposed machine learning paradigm, data programming.
We present a flexible interface layer for writing labeling functions based on our experience over the past year collaborating with companies, agencies, and research labs.
In a user study, subject matter experts build models $\userstudyspeedup\times$ faster and increase predictive performance an average $\userstudyimprovement\%$ versus seven hours of hand labeling.
We study the modeling tradeoffs in this new setting and propose an optimizer for automating tradeoff decisions that gives up to $\optimizerGMmaxspeedup\times$ speedup per pipeline execution.
In two collaborations, with the U.S. Department of Veterans Affairs and the U.S. Food and Drug Administration, and on four open-source text and image data sets representative of other deployments, Snorkel provides $\avgimprovementoverheuristic\%$ average improvements to predictive performance over prior heuristic approaches and comes within an average $\avgHLgap\%$ of the predictive performance of large hand-curated training sets.
\end{abstract}

%% file: intro.tex

\section{Introduction}
In the last several years, there has been an explosion of interest in machine-learning-based systems across industry, government, and academia, with an estimated spend this year of \$12.5 billion~\cite{idc17}.
A central driver has been the advent of \textit{deep learning} techniques, which can learn task-specific representations of input data, obviating what used to be the most time-consuming development task: feature engineering.
These learned representations are particularly effective for tasks like natural language processing and image analysis, which have high-dimensional, high-variance input that is impossible to fully capture with simple rules or hand-engineered features~\cite{graves2005framewise, deng2009imagenet}.
However, deep learning has a major upfront cost: these methods need massive \textit{training sets} of labeled examples to learn from---often tens of thousands to millions to reach peak predictive performance~\cite{sun2017revisiting}.

\begin{figure}[t]
\centering
\includegraphics[width=0.45\textwidth]{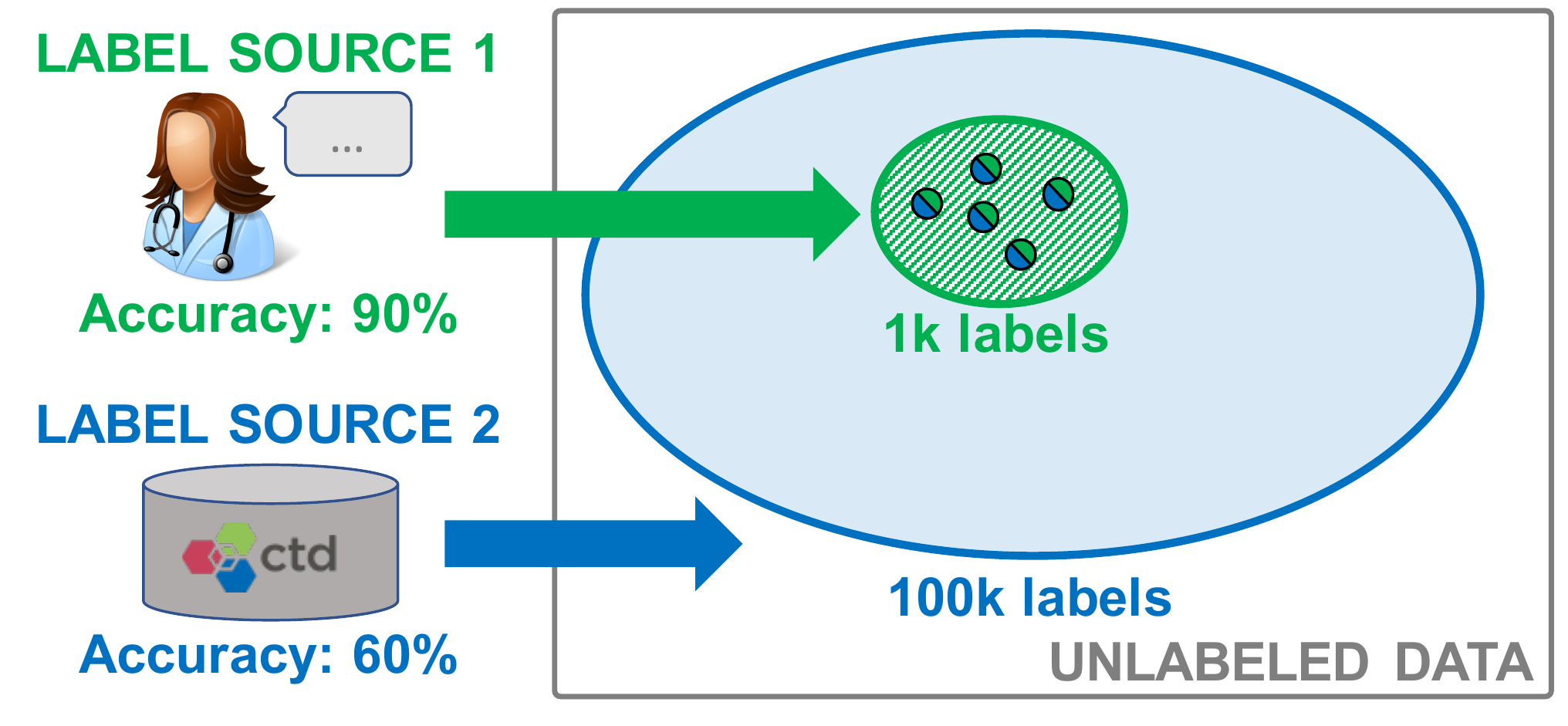}
\vspace{-10pt}
\caption{
In Example~\ref{ex:training-pt-ex}, training data is labeled by sources of differing accuracy and coverage.
Two key challenges arise in using this weak supervision effectively.
First, we need a way to estimate the unknown source accuracies to resolve disagreements. 
Second, we need to pass on this critical lineage information to the end model being trained.
}
\vspace{-2em}
\label{fig:training-pt-ex}
\end{figure}

Such training sets are enormously expensive to create, especially when domain expertise is required.
For example, reading scientific papers, analyzing intelligence data, and interpreting medical images all require labeling by trained \emph{subject matter experts} (SMEs).
Moreover, we observe from our engagements with collaborators like research labs and major technology companies that modeling goals such as class definitions or granularity change as projects progress, necessitating re-labeling.
Some big companies are able to absorb this cost, hiring large teams to label training data~\cite{wired16, time17, davis2013ctd}.
However, the bulk of practitioners are increasingly turning to \emph{weak supervision}: cheaper sources of labels that are noisier or heuristic.
The most popular form is \textit{distant supervision}, in which the records of an external knowledge base are heuristically aligned with data points to produce noisy labels~\cite{bunescu:acl07,mintz:acl09,alfonseca:acl12}.
Other forms include crowdsourced labels~\cite{yuen:socialcom11, quinn:chi11}, rules and heuristics for labeling data~\cite{zhang:cacm17, rekatsinas:vldb17}, and others~\cite{zaidan:emnlp08, liang:icml09, mann:jmlr10, mann:jmlr10, stewart:aaai17}.
While these sources are inexpensive, they often have limited accuracy and coverage.

Ideally, we would combine the labels from many weak supervision sources to increase the accuracy and coverage of our training set.
However, two key challenges arise in doing so effectively.
First, sources will overlap and conflict, and to resolve their conflicts we need to estimate their accuracies and correlation structure, \emph{without} access to ground truth.
Second, we need to pass on critical lineage information about label quality to the end model being trained.

\vspace{-5pt}
\begin{example}
\label{ex:training-pt-ex}
In Figure~\ref{fig:training-pt-ex}, we obtain labels from a high accuracy, low coverage Source 1, and from a low accuracy, high coverage Source 2, which overlap and disagree (split-color points).
If we take an unweighted majority vote to resolve conflicts, we end up with null (tie-vote) labels.
If we could correctly estimate the source accuracies, we would resolve conflicts in the direction of Source 1.

We would still need to pass this information on to the end model being trained.
Suppose that we took labels from Source 1 where available, and otherwise took labels from Source 2.
Then, the expected training set accuracy would be $60.3\%$---only marginally better than the weaker source.
Instead we should represent training label lineage in end model training, weighting labels generated by high-accuracy sources more.
\end{example}
\vskip -1em

In recent work, we developed \emph{data programming} as a \linebreak paradigm for addressing both of these challenges by modeling multiple label sources without access to ground truth, and generating \textit{probabilistic} training labels representing the lineage of the individual labels.
We prove that, surprisingly, we can recover source accuracy and correlation structure without hand-labeled training data~\cite{ratner:nips16, bach:icml17}.
However, there are many practical aspects of implementing and applying this abstraction that have not been previously considered.

We present \emph{Snorkel}, the first end-to-end system for combining weak supervision sources to rapidly create training data.
We built Snorkel as a prototype to study how people could use data programming, a fundamentally new approach to building machine learning applications.
Through weekly hackathons and office hours held at Stanford University over the past year, we have interacted with a growing user community around Snorkel's open source implementation.\footnote{\small{\url{http://snorkel.stanford.edu}}}
We have observed SMEs in industry, science, and government deploying Snorkel for knowledge base construction, image analysis, bioinformatics, fraud detection, and more.
From this experience, we have distilled three principles that have shaped Snorkel's design:
\begin{enumerate} \compactify
\item {\bf Bring All Sources to Bear:} The system should enable users to opportunistically use labels from all available weak supervision sources.

\item {\bf Training Data as the Interface to ML:} The system should model label sources to produce a single, probabilistic label for each data point and train any of a wide range of classifiers to generalize beyond those sources.

\item  {\bf Supervision as Interactive Programming:} The system should provide rapid results in response to user supervision. We envision weak supervision as the REPL-like interface for machine learning.
\end{enumerate}

\begin{figure*}[t]
\begin{center}
\centerline{\includegraphics[width=2\columnwidth]{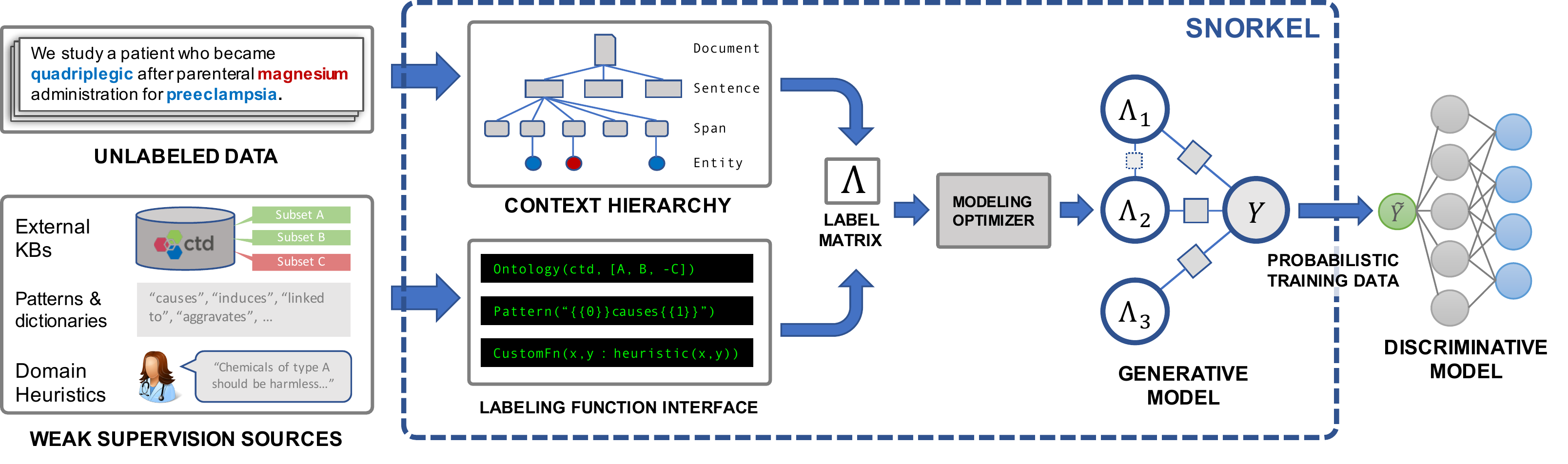}}
\vskip -.1in
\caption{An overview of the Snorkel system. (1) SME users write \emph{labeling functions (LFs)} that express weak supervision sources like distant supervision, patterns, and heuristics.
(2) Snorkel applies the LFs over unlabeled data and learns a generative model to combine the LFs' outputs into probabilistic labels. (3) Snorkel uses these labels to train a discriminative classification model, such as a deep neural network.}
\label{fig:system}
\end{center}
\vskip -0.4in
\end{figure*}

\noindent Our work makes the following technical contributions:
\vspace{-.5em} \\ \\
{\bf A Flexible Interface for Sources:}
We observe that the heterogeneity of weak supervision strategies is a stumbling block for developers.
    Different types of weak supervision operate on different scopes of the input data.
    For example, distant supervision has to be mapped programmatically to specific spans of text.
    Crowd workers and weak classifiers often operate over entire documents or images.
    Heuristic rules are open ended; they can leverage information from multiple contexts simultaneously, such as combining information from a document's title, named entities in the text, and knowledge bases.
    This heterogeneity was cumbersome enough to completely block users of early versions of Snorkel.

To address this challenge, we built an interface layer \linebreak around the abstract concept of a \textit{labeling function (LF)}.
We developed a flexible language for expressing weak supervision strategies and supporting data structures.
We observed accelerated user productivity with these tools, which we validated in a user study where SMEs build models $\userstudyspeedup \times$ faster and increase predictive performance an average $\userstudyimprovement\%$ versus seven hours of hand labeling.
\vspace{-.5em} \\ \\
\noindent {\bf Tradeoffs in Modeling of Sources:}
Snorkel learns the accuracies of weak supervision sources without access to ground truth using a generative model \cite{ratner:nips16}.
Furthermore, it also learns correlations and other statistical dependencies among sources, correcting for dependencies in labeling functions that skew the estimated accuracies \cite{bach:icml17}.
This paradigm gives rise to previously unexplored tradeoff spaces between predictive performance and speed.
The natural first question is: when does modeling the accuracies of sources improve predictive performance?
Further, how many dependencies, such as correlations, are worth modeling?

We study the tradeoffs between predictive performance and training time in generative models for weak supervision.
While modeling source accuracies and correlations will not hurt predictive performance, we present a theoretical analysis of when a simple majority vote will work just as well.
Based on our conclusions, we introduce an optimizer for deciding when to model accuracies of labeling functions, and when learning can be skipped in favor of a simple majority vote.
Further, our optimizer automatically decides which correlations to model among labeling functions.
This optimizer correctly predicts the advantage of generative modeling over majority vote to within $\optimizerGMadvacc$ accuracy points on average on our evaluation tasks, and accelerates pipeline executions by up to $\optimizerGMmaxspeedup\times$.
It also enables us to gain 60\%--70\% of the benefit of correlation learning while saving up to $\optimizerSLmaxtimesaved\%$ of training time ($\optimizerSLmaxtimesavedmin$ minutes per execution).
\vspace{-.5em} \\ \\
\noindent {\bf First End-to-End System for Data Programming:} Snorkel is the first system to implement our recent work on data programming \cite{ratner:nips16, bach:icml17}.
Previous ML systems that we and others developed \cite{zhang:cacm17} required extensive feature engineering and model specification, leading to confusion about where to inject relevant domain knowledge.
    While programming weak supervision seems superficially similar to feature engineering, we observe that users approach the two processes very differently.
    Our vision---weak supervision as the sole port of interaction for machine learning---implies radically different workflows, requiring a proof of concept.

Snorkel demonstrates that this paradigm enables users to develop high-quality models for a wide range of tasks.
We report on two deployments of Snorkel, in collaboration with the U.S. Department of Veterans Affairs and Stanford Hospital and Clinics, and the U.S. Food and Drug Administration, where Snorkel improves over heuristic baselines by an average $\avgimprovementoverheuristicRW\%$.
We also report results on four open-source datasets that are representative of other Snorkel deployments, including bioinformatics, medical image analysis, and crowdsourcing; on which Snorkel beats heuristics by an average $\avgimprovementoverheuristicBM\%$ and comes within an average $\avgHLgap\%$ of the predictive performance of large hand-curated training sets.
\vspace{-0.5em}

%% file: system.tex

\section{Snorkel Architecture}
\label{sec:system}

Snorkel's workflow is designed around data programming \cite{ratner:nips16,bach:icml17}, a fundamentally new paradigm for training machine learning models using weak supervision, and proceeds in three main stages (Figure~\ref{fig:system}):
\begin{enumerate}\compactify
    \item \textbf{Writing Labeling Functions:} Rather than \linebreak hand-labeling training data, users of Snorkel write labeling functions, which allow them to express various weak supervision sources such as patterns, heuristics, external knowledge bases, and more.
    This was the component most informed by early interactions (and mistakes) with users over the last year of deployment, and we present a flexible interface and supporting data model.
    
    \item \textbf{Modeling Accuracies and Correlations:} Next, \linebreak Snorkel automatically learns a \textit{generative model} over the labeling functions, which allows it to estimate their accuracies and correlations.
    This step uses no ground-truth data, learning instead from the agreements and disagreements of the labeling functions.
    We observe that this step improves end predictive performance $\avgrelimprovementoverMV\%$ over Snorkel with unweighted label combination, and anecdotally that it streamlines the user development experience by providing actionable feedback about labeling function quality.
    
    \item \textbf{Training a Discriminative Model:} The output of Snorkel is a set of \textit{probabilistic labels} that can be used to train a wide variety of state-of-the-art machine learning models, such as popular deep learning models.
    While the generative model is essentially a re-weighted combination of the user-provided labeling functions---which tend to be precise but low-coverage---modern discriminative models can retain this precision while learning to generalize beyond the labeling functions, increasing coverage and robustness on unseen data.
\end{enumerate}

Next we set up the problem Snorkel addresses and describe its main components and design decisions.
\\ \\ \\
\noindent {\bf Setup:}
Our goal is to learn a parameterized classification model $h_\theta$ that, given a data point $\x \in \domx$, predicts its label $\y \in \domy$, where the set of possible labels $\domy$ is discrete.
For simplicity, we focus on the binary setting $\domy = \{-1, 1\}$, though we include a multi-class application in our experiments.
For example, $x$ might be a medical image, and $y$ a label indicating normal versus abnormal.
In the relation extraction examples we look at, we often refer to $x$ as a \textit{candidate}.
In a traditional supervised learning setup, we would learn $\discmodel_\discparam$ by fitting it to a \textit{training set} of labeled data points.
However, in our setting, we assume that we only have access to unlabeled data for training.
We do assume access to a small set of labeled data used during development, called the \textit{development set}, and a blind, held-out labeled \textit{test set} for evaluation.
These sets can be orders of magnitudes smaller than a training set, making them economical to obtain.

The user of Snorkel aims to generate training labels by providing a set of labeling functions, which are black-box functions, $\lf : \domx \rightarrow \domy \cup \{\emptyset\}$,  that take in a data point and output a label
where we use $\emptyset$ to denote that the labeling functions abstains.
Given $\ny$ unlabeled data points and $\nlf$ labeling functions, Snorkel applies the labeling functions over the unlabeled data to produce a matrix of labeling function outputs $\lfout \in \left(\domy \cup \{ \emptyset \} \right)^{\ny \times \nlf}$.
The goal of the remaining Snorkel pipeline is to synthesize this label matrix $\lfout$---which may contain overlapping and conflicting labels for each data point---into a single vector of \textit{probabilistic training labels} $\tilde{Y} = (\tilde{\y}_{1}, ..., \tilde{\y}_{\ny})$, where $\tilde{y}_\iy \in [0,1]$.
These training labels can then be used to train a discriminative model.

Next, we introduce the running example of a text relation extraction task as a proxy for many real-world knowledge base construction and data analysis tasks:

\begin{example}
Consider the task of extracting mentions of adverse chemical-disease relations from the biomedical literature (see CDR task, Section~\ref{sec:applications}).
Given documents with mentions of chemicals and diseases tagged, we refer to each co-occuring (chemical, disease) mention pair as a \textit{candidate} extraction, which we view as a data point to be classified as either true or false.
For example, in Figure~\ref{sec:system}, we would have two candidates with true labels \texttt{$y_1$ = True} and \texttt{$y_2$ = False}:
\begin{python}
x_1 = Causes("magnesium", "quadriplegic")
x_2 = Causes("magnesium", "preeclampsia")
\end{python}
\end{example}

\begin{figure}[t]
\centering
\includegraphics[width=0.45\textwidth]{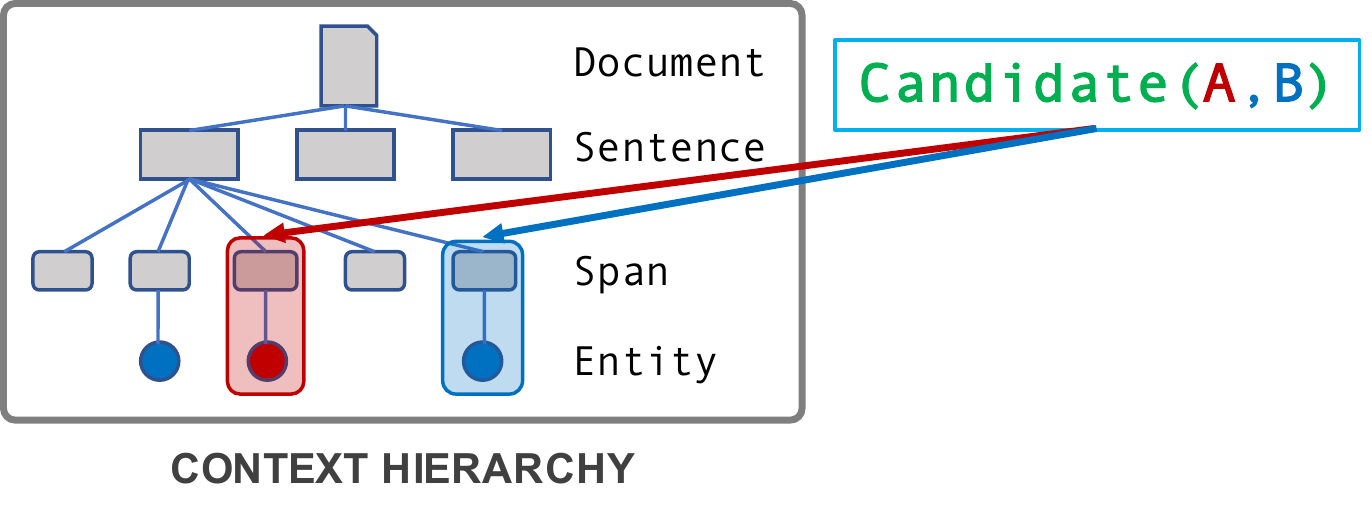}
\vspace{-10pt}
\caption[...]{
  Labeling functions take as input a \texttt{Candidate} object, representing a data point to be classified.
  Each \texttt{Candidate} is a tuple of \texttt{Context} objects, which are part of a hierarchy representing the local context of the \texttt{Candidate}.
}
\vspace{-10pt}
\label{fig:candidate-context}
\end{figure}

\noindent {\bf Data Model:}
A design challenge is managing complex, unstructured data in a way that enables SMEs to write labeling functions over it.
In Snorkel, input data is stored in a \emph{context hierarchy}.
It is made up of context types connected by parent/child relationships, which are stored in a relational database and made available via an object-relational mapping (ORM) layer built with SQLAlchemy.\footnote{\small{\url{https://www.sqlalchemy.org/}}}
Each \emph{context type} represents a conceptual component of data to be processed by the system or used when writing labeling functions; for example a document, an image, a paragraph, a sentence, or an embedded table.
Candidates---i.e., data points $x$---are then defined as tuples of contexts (Figure~\ref{fig:candidate-context}).

\begin{example}
In our running CDR example, the input documents can be represented in Snorkel as a hierarchy consisting of \texttt{Documents}, each containing one or more \linebreak \texttt{Sentences}, each containing one or more \texttt{Spans} of text. These \texttt{Spans} may also be tagged with metadata, such as \texttt{Entity} markers identifying them as chemical or disease mentions (Figure~\ref{fig:candidate-context}).
A candidate is then a tuple of two \texttt{Spans}.
\end{example}

\subsection{A Language for Weak Supervision}
\label{sec:lfs}

Snorkel uses the core abstraction of a labeling function to allow users to specify a wide range of weak supervision sources such as patterns, heuristics, external knowledge bases, crowdsourced labels, and more.
This higher-level, less precise input is more efficient to provide (see Section~\ref{sec:user_study}),
and can be automatically denoised and synthesized, as described in subsequent sections.

In this section, we describe our design choices in building an interface for writing labeling functions, which we envision as a unifying programming language for weak supervision.
These choices were informed to a large degree by our interactions---primarily through weekly office hours---with Snorkel users in bioinformatics, defense, industry, and other areas over the past year.\footnote{\small{\url{http://snorkel.stanford.edu\#users}}}
For example, while we initially intended to have a more complex structure for labeling functions, with manually specified types and correlation structure, we quickly found that simplicity in this respect was critical to usability (and not empirically detrimental to our ability to model their outputs).
We also quickly discovered that users wanted either far more expressivity or far less of it, compared to our first library of function templates.
We thus trade off expressivity and efficiency by allowing users to write labeling functions at two levels of abstraction: custom Python functions and declarative operators.
\\ \\
\noindent {\bf Hand-Defined Labeling Functions:}
In its most general form, a labeling function is just an arbitrary snippet of code, usually written in Python, which accepts as input a \texttt{Candidate} object and either outputs a label or abstains.
Often these functions are similar to extract-transform-load scripts, expressing basic patterns or heuristics, but may use supporting code or resources and be arbitrarily complex.
Writing labeling functions by hand is supported by the ORM layer, which maps the context hierarchy and associated metadata to an object-oriented syntax, allowing the user to easily traverse the structure of the input data.

\begin{example}\label{ex:cdr_lfs}
In our running example, we can write a labeling function that checks if the word ``causes" appears between the chemical and disease mentions.
If it does, it outputs \texttt{True} if the chemical mention is first and \texttt{False} if the disease mention is first.
If ``causes'' does not appear, it outputs \texttt{None}, indicating abstention:
\begin{python}
def LF_causes(x):
  cs, ce = x.chemical.get_word_range()
  ds, de = x.disease.get_word_range()
  if ce < ds and "causes" in x.parent.words[ce+1:ds]:
      return True
  if de < cs and "causes" in x.parent.words[de+1:cs]:
      return False
  return None
\end{python}
We could also write this with Snorkel's declarative interface:
\begin{python}
LF_causes = lf_search("{{1}}.*\Wcauses\W.*{{2}}", reverse_args=False)
\end{python}
\end{example}
\noindent {\bf Declarative Labeling Functions:}
\systemx~includes a library of declarative operators that encode the most common weak supervision function types, based on our experience with users over the last year.
These functions capture a range of common forms of weak supervision, for example:
\begin{itemize} \compactify
    \item \textbf{Pattern-based:}
    Pattern-based heuristics embody the motivation of soliciting higher information density input from SMEs.
    For example, pattern-based heuristics encompass feature annotations~\cite{zaidan:emnlp08} and pattern-bootstrapping approaches~\cite{hearst1992automatic,gupta2014improved} (Example~\ref{ex:cdr_lfs}).
    
    \item \textbf{Distant supervision:}
    Distant supervision generates training labels by heuristically aligning data points with an external knowledge base, and is one of the most popular forms of weak supervision~\cite{mintz:acl09,alfonseca:acl12,hoffmann:acl11}.
    
    \item \textbf{Weak classifiers:}
    Classifiers that are insufficient for our task---e.g., limited coverage, noisy, biased, and/or trained on a different dataset---can be used as labeling functions.
    
    \item \textbf{Labeling function generators:} One higher-level abstraction that we can build on top of labeling functions in \systemx~is \textit{labeling function generators}, which generate multiple labeling functions from a single resource, such as crowdsourced labels and distant supervision from structured knowledge bases (Example~\ref{ex:generator}).

\end{itemize}

\begin{example}
A challenge in traditional distant supervision is that different subsets of knowledge bases have different levels of accuracy and coverage.
In our running example, we can use the Comparative Toxicogenomics Database (CTD)\footnote{\small{\url{http://ctdbase.org/}}} as distant supervision, separately modeling different subsets of it with separate labeling functions.
For example, we might write one labeling function to label a candidate \texttt{True} if it occurs in the ``Causes'' subset, and another to label it \texttt{False} if it occurs in the ``Treats'' subset.
We can write this using a labeling function generator,
\begin{python}
LFs_CTD = Ontology(ctd,
    {"Causes": True, "Treats": False})
\end{python}
which creates two labeling functions.
In this way, generators can be connected to large resources and create hundreds of labeling functions with a line of code.
\label{ex:generator}
\end{example}

\subsection{Generative Model}

The core operation of \systemx~is modeling and integrating the noisy signals provided by a set of labeling functions.
Using the recently proposed approach of data programming~\cite{ratner:nips16, bach:icml17}, we model the true class label for a data point as a latent variable in a probabilistic model.
In the simplest case, we model each labeling function as a noisy ``voter'' which is \emph{independent}---i.e., makes errors that are uncorrelated with the other labeling functions.
This defines a generative model of the votes of the labeling functions as noisy signals about the true label.

We can also model statistical dependencies between the labeling functions to improve predictive performance.
For example, if two labeling functions express similar heuristics, we can include this dependency in the model and avoid a ``double counting'' problem.
We observe that such pairwise correlations are the most common, so we focus on them in this paper (though handling higher order dependencies is straightforward).
We use our structure learning method for generative models \cite{bach:icml17} to select a set $C$ of labeling function pairs $(\ilf, \ilfalt)$ to model as correlated (see Section~\ref{sec:sl_optimizer}).

Now we can construct the full generative model as a factor graph.
We first apply all the labeling functions to the unlabeled data points, resulting in a label matrix $\lfout$, where $\lfout_{\iy, \ilf} = \lf_\ilf(\x_\iy)$.
We then encode the generative model $\p_{\genparam}(\lfout, \Y)$ using three factor types, representing the labeling propensity, accuracy, and pairwise correlations of labeling functions:
\begin{align*}
\dep_{\iy, \ilf}^{\textrm{Lab}}(\lfout, \Y) &= \mathds{1}\{\lfout_{\iy, \ilf} \neq \emptyset\} & \\
\dep_{\iy, \ilf}^{\textrm{Acc}}(\lfout, \Y) &= \mathds{1}\{\lfout_{\iy, \ilf} = \y_\iy\} &  \\
\dep_{\iy, \ilf, \ilfalt}^{\textrm{Corr}}(\lfout, \Y) &= \mathds{1}\{\lfout_{\iy, \ilf} = \lfout_{\iy, \ilfalt}\} & (\ilf, \ilfalt)\in C
\end{align*}
For a given data point $\x_\iy$, we define the concatenated vector of these factors for all the labeling functions $\ilf=1,...,\nlf$ and potential correlations $C$ as $\dep_\iy(\lfout, \Y)$, and the corresponding vector of parameters $\genparam \in \mathbb{R}^{2\nlf + |C|}$.
This defines our model:
\begin{align*}
p_{\genparam}(\lfout, \Y) &= Z_{\genparam}^{-1}\exp\left(\sum_{\iy=1}^\ny \genparam^T \dep_\iy(\lfout, \y_\iy)\right)~,
\end{align*}
where $Z_\genparam$ is a normalizing constant.
To learn this model \textit{without} access to the true labels $\Y$, we minimize the negative log marginal likelihood given the observed label matrix $\lfout$:
\begin{align*}
\hat{\genparam} = \argmin_{\genparam}-\log\sum_\Y \p_{\genparam}(\lfout, \Y)~.
\end{align*}

We optimize this objective by interleaving stochastic gradient descent steps with Gibbs sampling ones, similar to contrastive divergence~\cite{hinton2002training}; for more details, see~\cite{ratner:nips16,bach:icml17}.
We use the Numbskull library,\footnote{\url{https://github.com/HazyResearch/numbskull}} a Python NUMBA-based Gibbs sampler.
We then use the predictions, $\probY = \p_{\hat{\genparam}}(\Y | \lfout)$, as \textit{probabilistic training labels}.

\subsection{Discriminative Model}

The end goal in Snorkel is to train a model that generalizes beyond the information expressed in the labeling functions.
We train a discriminative model $h_\discparam$ on our probabilistic labels $\tilde{Y}$ by minimizing a \textit{noise-aware} variant of the loss  $l(\discmodel_\discparam(\x_\iy), \y)$, i.e., the expected loss with respect to $\tilde{Y}$:
\[
\hat{\discparam} = \argmin_\discparam \sum_{\iy = 1}^\ny \mathbb{E}_{\y \sim \tilde{Y}} \left[
  l(\discmodel_\discparam(\x_\iy), \y)
\right].
\]

A formal analysis shows that as we increase the amount of \emph{unlabeled data}, the generalization error of discriminative models trained with Snorkel will decrease at the same asymptotic rate as traditional supervised learning models do with additional hand-labeled data \cite{ratner:nips16},
allowing us to increase predictive performance by adding more unlabeled data.
Intuitively, this property holds because as more data is provided, the discriminative model sees more features that co-occur with the heuristics encoded in the labeling functions.

\begin{example}
The CDR data contains the sentence, \linebreak ``Myasthenia gravis presenting as weakness after magnesium administration.''
None of the 33 labeling functions we developed vote on the corresponding \texttt{Causes(magnesium,} \linebreak \texttt{myasthenia gravis)} candidate, i.e., they all abstain.
However, a deep neural network trained on probabilistic training labels from Snorkel correctly identifies it as a true mention.
\label{ex:generalization}
\end{example}

Snorkel provides connectors for popular machine learning libraries such as TensorFlow \cite{abadi:osdi16}, allowing users to exploit commodity models like deep neural networks that do not require hand-engineering of features and have robust predictive performance across a wide range of tasks.

%% file: modeling.tex

\section{Weak Supervision Tradeoffs}
\label{sec:modeling}

We study the fundamental question of when---and at what level of complexity---we should expect Snorkel's generative model to yield the greatest predictive performance gains.
Understanding these performance regimes can help guide users, and introduces a tradeoff space between predictive performance and speed.
We characterize this space in two parts: first, by analyzing when the generative model can be approximated by an unweighted majority vote, and second, by automatically selecting the complexity of the correlation structure to model.
We then introduce a two-stage, rule-based optimizer to support fast development cycles.

\input{mv_vs_gm_optimizer}

\input{sl_optimizer}

%% file: mv_vs_gm_optimizer.tex

\subsection{Modeling Accuracies}

The natural first question when studying systems for weak supervision is, ``When does modeling the accuracies of \linebreak sources improve end-to-end predictive performance?''
We study that question in this subsection and propose a heuristic to identify settings in which this modeling step is most beneficial.

\subsubsection{Tradeoff Space}
We start by considering the \textit{label density} $d_\lfout$ of the label matrix $\lfout$, defined as the mean number of non-abstention labels per data point.
In the low-density setting, sparsity of labels will mean that there is limited room for even an optimal weighting of the labeling functions to diverge much from the majority vote.
Conversely, as the label density grows, known theory confirms that the majority vote will eventually be optimal \cite{li2013error}.
It is the middle-density regime where we expect to most benefit from applying the generative model.
We start by defining a measure of the benefit of weighting the labeling functions by their true accuracies---in other words, the predictions of a perfectly estimated generative model---versus an unweighted majority vote:
\begin{definition}{\textbf{(Modeling Advantage)}}
Let the weighted majority vote of $\nlf$ labeling functions on data point $x_\iy$ be denoted as $f_\genparam(\lfout_{\iy}) = \sum_{\ilf=1}^\nlf \genparam_\ilf\lfout_{\iy,\ilf}$, and the unweighted majority vote (MV) as $f_1(\lfout_{\iy}) = \sum_{\ilf=1}^\nlf \lfout_{\iy,\ilf}$, where we consider the binary classification setting and represent an abstaining vote as $0$.
We define the \textbf{modeling advantage} $A_\genparam$ as the improvement in accuracy of $f_\genparam$ over $f_1$ for a dataset:
\begin{align*}
A_\genparam(\lfout, y) = 
  \frac{1}{\ny}\sum_{\iy=1}^\ny\left(
        \mathbbm{1}\left\{
            y_\iy f_\genparam(\lfout_\iy) > 0 \wedge y_\iy f_1(\lfout_\iy) \leq 0
        \right\}\right.\\
       -\left.\mathbbm{1}\left\{y_\iy f_\genparam(\lfout_\iy) \leq 0 \wedge y_\iy f_1(\lfout_\iy) > 0 \right\}\right)
\end{align*}
In other words, $A_\genparam$ is the number of times $f_\genparam$ correctly disagrees with $f_1$ on a label, minus the number of times it incorrectly disagrees.
Let the \textbf{optimal advantage} $A^* = A_{\genparam^*}$ be the advantage using the optimal weights $\genparam^*$ (WMV*).
\end{definition}

\begin{figure}[t]
\centering
\includegraphics[width=0.5\textwidth]{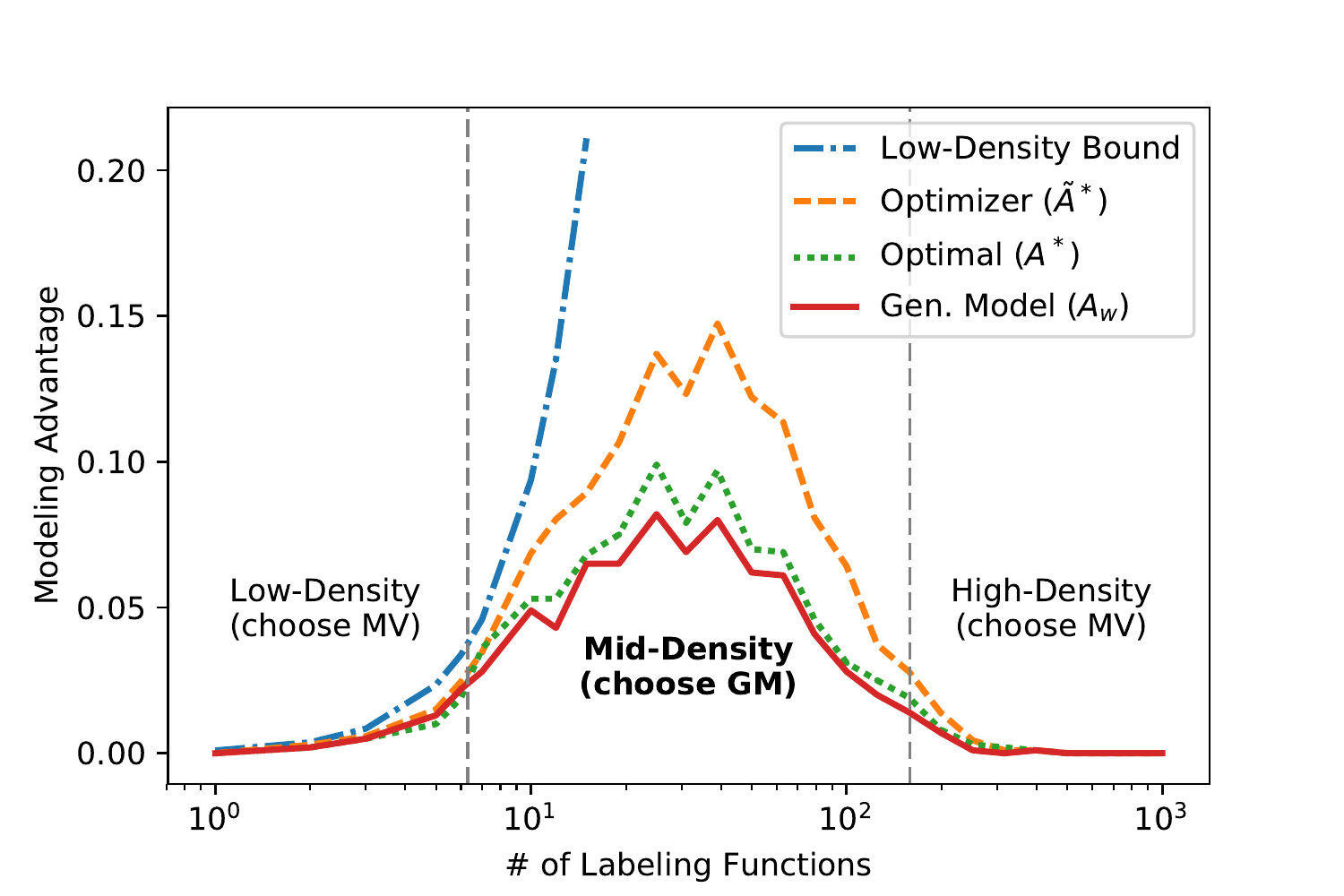}
\vspace{-20pt}
\caption[...]{
A plot of the \textit{modeling advantage}, i.e., the improvement in label accuracy from the generative  model, as a function of the number of labeling functions (equivalently, the label density) on a synthetic dataset.\footnotemark~
We plot the advantage obtained by a learned generative model (GM), $A_\genparam$; by an optimal model $A^*$; the upper bound $\tilde{A}^*$ used in our optimizer; and the low-density bound (Proposition~\ref{thm:low-density}).
}
\vspace{-10pt}
\label{fig:label-density-regimes}
\end{figure}

\footnotetext{\small{We generate a class-balanced dataset of $\ny=1000$ data points with binary labels, and $\nlf$ independent labeling functions with average accuracy 75\% and a fixed 10\% probability of voting.}}

To build intuition, we start by analyzing the optimal advantage for three regimes of label density (see Figure~\ref{fig:label-density-regimes}):
\\ \\
\noindent {\bf Low Label Density:}
In this sparse setting, very few data points have more than one non-abstaining label; only a small number have multiple conflicting labels.
We have observed this occurring, for example, in the early stages of application development.
We see that with non-adversarial labeling functions ($\genparam^*>0$), even an optimal generative model (WMV*) can only disagree with MV when there are disagreeing labels, which will occur infrequently.
We see that the expected optimal advantage will have an upper bound that falls quadratically with label density:

\begin{proposition}{\textbf{(Low-Density Upper Bound)}}
\label{thm:low-density}
Assume that $P(\lfout_{\iy,\ilf} \neq 0) = p_l\ \forall \iy,\ilf$, and $\genparam^*_\ilf>0\ \forall \ilf$.
Then, the expected label density is $\bar{d} = \nlf p_l$, and
\begin{align}
\mathbb{E}_{\lfout,y,\genparam^*}\left[A^*\right] = O\left(\bar{d}^2\right)
\end{align}
\end{proposition}
\textit{Proof Sketch:} We bound the advantage above by computing the expected number of pairwise disagreements.
\\ \\
\noindent {\bf High Label Density:}
In this setting, the majority of the data points have a large number of labels. For example, we might be working in an extremely high-volume crowdsourcing setting, or an application with many high-coverage knowledge bases as distant supervision. Under modest assumptions---namely, that the average labeling function accuracy $\overline{\alpha}^*$ is greater than 50\%---it is known that the majority vote converges exponentially to an optimal solution as the average label density $\bar{d}$ increases, which serves as an upper bound for the expected optimal advantage as well:

\begin{theorem}{\textit{(High-Density Upper Bound \cite{li2013error})}}
Assume that $P(\lfout_{\iy,\ilf}\neq 0) = p_l\ \forall \iy,\ilf$, and that $\overline{\alpha}^* = \frac{1}{\nlf}\sum_{\ilf=1}^\nlf \alpha^*_\ilf = \frac{1}{\nlf}\sum_{\ilf=1}^\nlf 1/(1+\exp(\genparam^*_\ilf)) > \frac12$.
Then:
\begin{align}
\mathbb{E}_{\lfout, y,\genparam^*}\left[A^*\right] \leq e^{-2p_l\left(\overline{\alpha}^*-\frac12\right)^2\bar{d}}
\end{align}
\end{theorem}
\textit{Proof:} This follows from the result in~\cite{li2013error} for the symmetric Dawid-Skene model under constant probability sampling.
\\ \\
\noindent {\bf Medium Label Density:}
In this middle regime, we expect that modeling the accuracies of the labeling functions will deliver the greatest gains in predictive performance because we will have many data points with a small number of disagreeing labeling functions.
For such points, the estimated labeling function accuracies can heavily affect the predicted labels.
We indeed see gains in the empirical results using an independent generative model that only includes accuracy factors $\dep_{\iy, \ilf}^{\textrm{Acc}}$ (Table~\ref{tab:mv_gm_opt_results}).
Furthermore, the guarantees in~\cite{ratner:nips16} establish that we can learn the optimal weights, and thus approach the optimal advantage.

\subsubsection{Automatically Choosing a Modeling Strategy}

\begin{table}
\centering
\caption{
Modeling advantage $A_\genparam$ attained using a generative model for several applications in Snorkel (Section~\ref{sec:applications}), the upper bound $\tilde{A}^*$ used by our optimizer, the modeling strategy selected by the optimizer---either majority vote (MV) or generative model (GM)---and the empirical label density $d_\lfout$.}
\label{tab:mv_gm_opt_results}
\begin{tabular}{l c c c c}
\toprule
Dataset & $A_\genparam$ (\%) & $\tilde{A}^*$ (\%) & Modeling Strategy & $d_\lfout$\\
\midrule
Radiology & 7.0 & 12.4 & \textbf{GM} & 2.3\\
CDR & 4.9 & 7.9 & \textbf{GM} & 1.8 \\
Spouses & 4.4 & 4.6 & \textbf{GM} & 1.4 \\
Chem & 0.1 & 0.3 & \textbf{MV} & 1.2 \\
EHR & 2.8 & 4.8 & \textbf{GM} & 1.2 \\
\bottomrule
\end{tabular}
\end{table}

The bounds in the previous subsection imply that there are settings in which we should be able to safely skip modeling the labeling function accuracies, simply taking the unweighted majority vote instead.
However, in practice, the overall label density $d_\lfout$ is insufficiently precise to determine the transition points of interest, given a user time-cost tradeoff preference (characterized by the \textit{advantage tolerance} parameter $\gamma$ in Algorithm~\ref{alg:modeling-optimizer}).
We show this in Table~\ref{tab:mv_gm_opt_results} using our application data sets from Section~\ref{sec:applications}.
For example, we see that the Chem and EHR label matrices have equivalent label densities; however, modeling the labeling function accuracies has a much greater effect for EHR than for Chem.

Instead of simply considering the average label density $d_\lfout$, we instead develop a best-case heuristic based on looking at the ratio of positive to negative labels for each data point.
This heuristic serves as an upper bound to the true expected advantage, and thus we can use it to determine when we can safely skip training the generative model (see Algorithm~\ref{alg:modeling-optimizer}).
Let $c_y(\lfout_\iy) = \sum_{\ilf=1}^\nlf \mathbbm{1}\left\{ \lfout_{\iy,\ilf} = y\right\}$ be the counts of labels of class $y$ for $x_\iy$, and assume that the true labeling function weights lie within a fixed range, $\genparam_\ilf\in [\genparam_{min},\genparam_{max}]$ and have a mean $\bar{\genparam}$.\footnote{\small{We fix these at defaults of $(\genparam_{min},\bar{\genparam},\genparam_{max}) = (0.5, 1.0, 1.5)$, which corresponds to assuming labeling functions have accuracies between 62\% and 82\%, and an average accuracy of 73\%.}}
Then, define:
\begin{align*}
&\Phi(\lfout_\iy, y) = \mathbbm{1}\left\{
    c_y(\lfout_\iy)\genparam_{max} > c_{-y}(\lfout_\iy)\genparam_{min}
  \right\}\\
&\tilde{A}^*(\lfout) = \frac{1}{\ny}\sum_{\iy=1}^\ny
  \smashoperator[r]{\sum_{y\in\pm 1}}
    \mathbbm{1}\left\{ yf_1(\lfout_\iy) \leq 0\right\}
    \Phi(\lfout_\iy, y)
    \sigma(2f_{\bar{\genparam}}(\lfout_\iy)y)
\end{align*}
where $\sigma(\cdot)$ is the sigmoid function, $f_{\bar{\genparam}}$ is majority vote with all weights set to the mean $\bar{\genparam}$, and $\tilde{A}^*(\lfout)$ is the predicted modeling advantage used by our optimizer.
Essentially, we are taking the expected counts of instances in which a \linebreak weighted majority vote could possibly flip the incorrect predictions of unweighted majority vote under best case conditions, which is an upper bound for the expected advantage:

\begin{proposition}{(\textbf{Optimizer Upper Bound})}
Assume \linebreak that the labeling functions have accuracy parameters (log-odds weights) $\genparam_\ilf \in [\genparam_{min}, \genparam_{max}]$, and have $\mathbb{E}[\genparam] = \bar{\genparam}$.
Then:
\begin{align}
\mathbb{E}_{y,\genparam^*}\left[A^*\ |\ \lfout \right] \leq \tilde{A}^*(\lfout)
\end{align}
\end{proposition}
\textit{Proof Sketch:} We upper-bound the modeling advantage by the expected number of instances in which WMV* is correct and MV is incorrect. We then upper-bound this by using the best-case probability of the weighted majority vote being correct given $(\genparam_{min}, \genparam_{max})$.
\\

We apply $\tilde{A}^*$ to a synthetic dataset and plot in Figure~\ref{fig:label-density-regimes}.
Next, we compute $\tilde{A}^*$ for the labeling matrices from experiments in Section~\ref{sec:applications}, and compare with the empirical advantage of the trained generative models (Table~\ref{tab:mv_gm_opt_results}).
We see that our approximate quantity $\tilde{A}^*$ serves as a correct guide in all cases for determining which modeling strategy to select, which for the mature applications reported on is indeed most often the generative model.
However, we see that while EHR and Chem have equivalent label densities, our optimizer correctly predicts that Chem can be modeled with majority vote, speeding up each pipeline execution by $\optimizerGMmaxspeedup\times$.
We find in our applications that the optimizer can save execution time especially during the initial stages of iterative development (see full version).
\\ \\ \\

%% file: sl_optimizer.tex

\subsection{Modeling Structure}
\label{sec:sl_optimizer}

\begin{figure*}[t]
\begin{center}
\centerline{\includegraphics[width=.66\columnwidth]{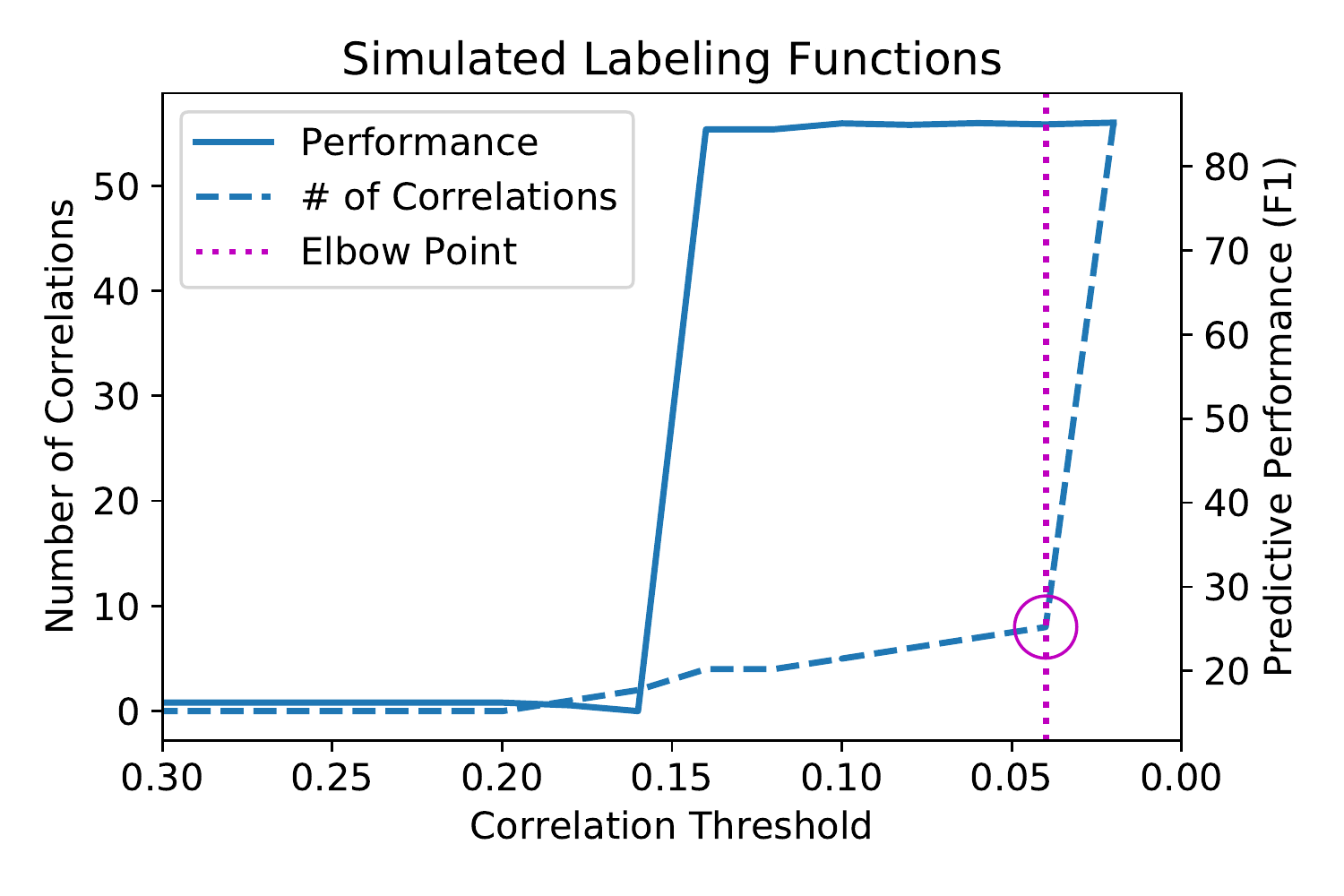}
\includegraphics[width=.66\columnwidth]{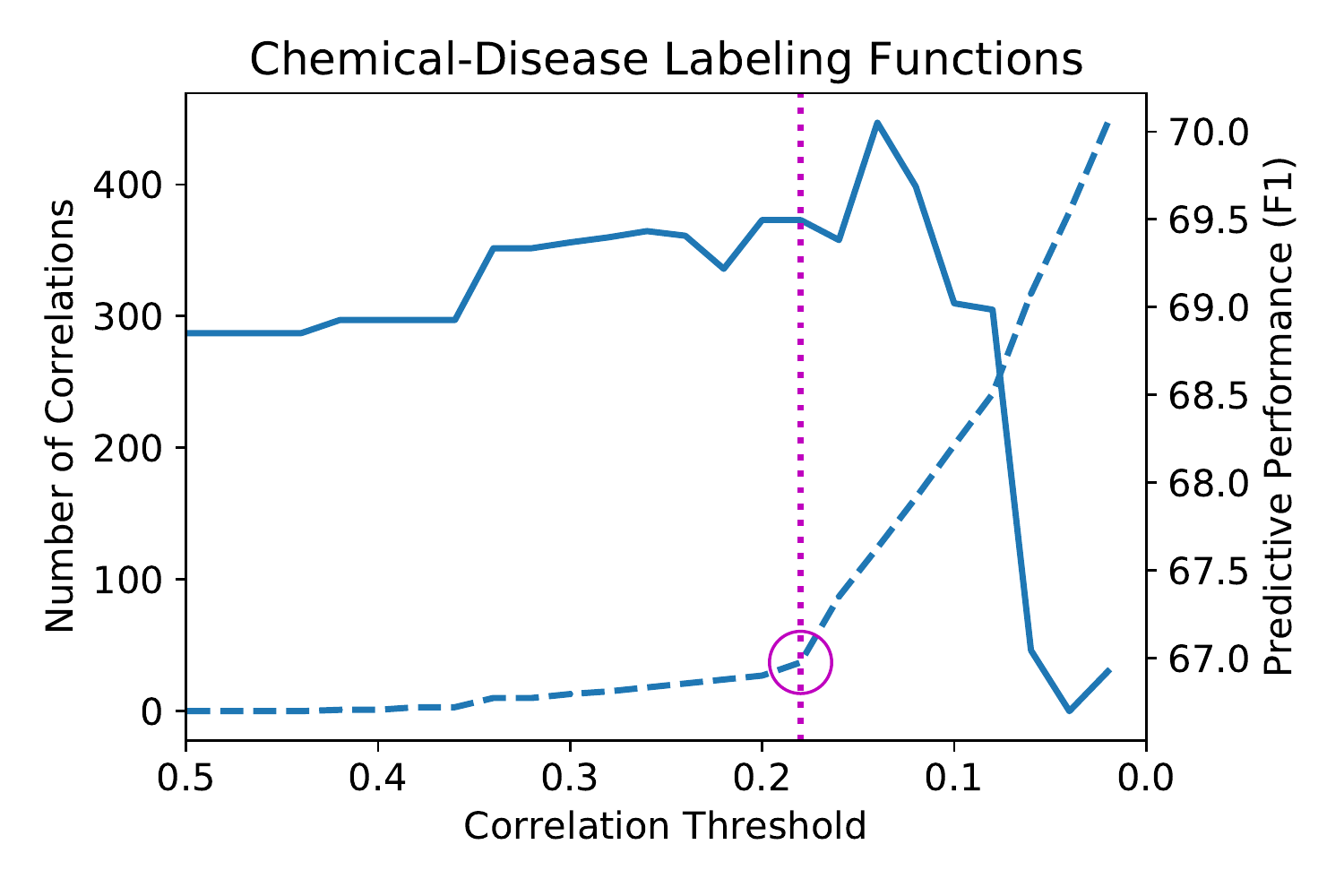}
\includegraphics[width=.66\columnwidth]{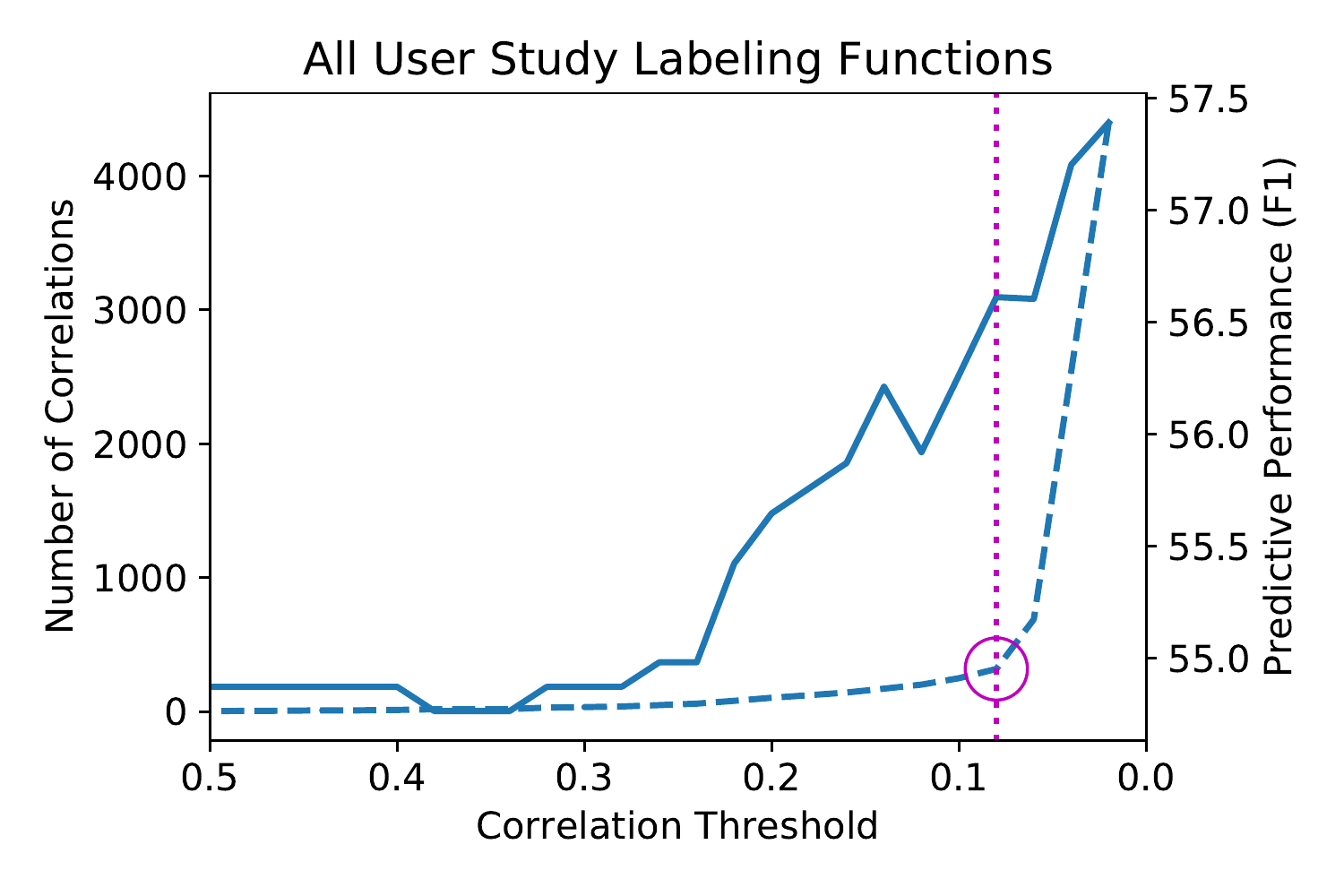}}
\vskip -0.1in
\caption{Predictive performance of the generative model and number of learned correlations versus the correlation threshold \boldmath$\threshold$.
The selected elbow point achieves a good tradeoff between predictive performance and computational cost (linear in the number of correlations).
Left: simulation of structure learning correcting the generative model.
Middle: the CDR task.
Right: all user study labeling functions for the Spouses task. }
\label{fig:sl_optimizer}
\end{center}
\vskip -0.4in
\end{figure*}

In this subsection, we consider modeling additional statistical structure beyond the independent model.
We study the tradeoff between predictive performance and computational cost, and describe how to automatically select a good point in this tradeoff space.

\paragraph*{Structure Learning}

We observe many Snorkel users writing labeling functions that are statistically dependent.
Examples we have observed include:
\begin{itemize} \compactify
\item Functions that are variations of each other, such as checking for matches against similar regular expressions.
\item Functions that operate on correlated inputs, such as raw tokens of text and their lemmatizations.
\item Functions that use correlated sources of knowledge, such as distant supervision from overlapping knowledge bases.
\end{itemize}
Modeling such dependencies is important because they affect our estimates of the true labels.
Consider the extreme case in which not accounting for dependencies is catastrophic:
\begin{example}
\label{ex:catastrophe}
Consider a set of 10 labeling functions, where 5 are perfectly correlated, i.e., they vote the same way on every data point, and 5 are conditionally independent given the true label.
If the correlated labeling functions have accuracy $\alpha = 50\%$ and the uncorrelated ones have accuracy $\beta = 99\%$,  then the maximum likelihood estimate of their accuracies according to the independent model is $\hat{\alpha} = 100\%$ and $\hat{\beta} = 50\%$.
\end{example}

Specifying a generative model to account for such dependencies by hand is impractical for three reasons.
First, it is difficult for non-expert users to specify these dependencies.
Second, as users iterate on their labeling functions, their dependency structure can change rapidly, like when a user relaxes a labeling function to label many more candidates.
Third, the dependency structure can be dataset specific, making it impossible to specify a priori, such as when a corpus contains many strings that match multiple regular expressions used in different labeling functions.
We observed users of earlier versions of Snorkel struggling for these reasons to construct accurate and efficient generative models with dependencies.
We therefore seek a method that can quickly identify an appropriate dependency structure from the labeling function outputs $\lfout$ alone.

Naively, we could include all dependencies of interest, such as all pairwise correlations, in the generative model and perform parameter estimation.
However, this approach is impractical.
For 100 labeling functions and 10,000 data points, estimating parameters with all possible correlations takes roughly 45 minutes.
When multiplied over repeated runs of hyperparameter searching and development cycles, this cost greatly inhibits labeling function development.
We therefore turn to our method for automatically selecting which dependencies to model without access to ground truth \cite{bach:icml17}.
It uses a pseudolikelihood estimator, which does not require any sampling or other approximations to compute the objective gradient exactly.
It is much faster than maximum likelihood estimation, taking 15 seconds to select pairwise correlations to be modeled among 100 labeling functions with 10,000 data points.
However, this approach relies on a selection threshold hyperparameter $\epsilon$ which induces a tradeoff space between predictive performance and computational cost.

\subsubsection{Tradeoff Space}
Such structure learning methods, whether pseudolikelihood or likelihood-based, crucially depend on a selection threshold $\epsilon$ for deciding which dependencies to add to the generative model.
Fundamentally, the choice of $\epsilon$ determines the complexity of the generative model.\footnote{\small{Specifically, $\epsilon$ is both the coefficient of the $\ell_1$ regularization term used to induce sparsity, and the minimum absolute weight in log scale that a dependency must have to be selected.}}
We study the tradeoff between predictive performance and computational cost that this induces.
We find that generally there is an ``elbow point'' beyond which the number of correlations selected---and thus the computational cost---explodes, and that this point is a safe tradeoff point between predictive performance and computation time.
\\ \\
\noindent {\bf Predictive Performance:}
At one extreme, a very large value of $\threshold$ will not include any correlations in the generative model, making it identical to the independent model.
As $\threshold$ is decreased, correlations will be added.
At first, when $\threshold$ is still high, only the strongest correlations will be included.
As these correlations are added, we observe that the generative model's predictive performance tends to improve.
Figure~\ref{fig:sl_optimizer}, left, shows the result of varying $\threshold$ in a simulation where more than half the labeling functions are correlated.
After adding a few key dependencies, the generative model resolves the discrepancies among the labeling functions.
Figure~\ref{fig:sl_optimizer}, middle, shows the effect of varying $\threshold$ for the CDR task.
Predictive performance improves as $\threshold$ decreases until the model overfits.
Finally, we consider a large number of labeling functions that are likely to be correlated.
In our user study (described in Section~\ref{sec:user_study}), participants wrote labeling functions for the Spouses task.
We combined all 125 of their functions and studied the effect of varying $\threshold$.
Here, we expect there to be many correlations since it is likely that users wrote redundant functions.
We see in Figure~\ref{fig:sl_optimizer}, right, that structure learning surpasses the best performing individual's generative model (50.0 F1).
\\ \\
\noindent {\bf Computational Cost:}
Computational cost is correlated with model complexity.
Since learning in Snorkel is done with a Gibbs sampler, the overhead of modeling additional correlations is linear in the number of correlations.
The dashed lines in Figure~\ref{fig:sl_optimizer} show the number of correlations included in each model versus $\threshold$.
For example, on the Spouses task, fitting the parameters of the generative model at $\threshold = 0.5$ takes 4 minutes, and fitting its parameters with $\threshold = 0.02$ takes 57 minutes.
Further, parameter estimation is often run repeatedly during development for two reasons: (i) fitting generative model hyperparameters using a development set requires repeated runs, and (ii) as users iterate on their labeling functions, they must re-estimate the generative model to evaluate them.

\subsubsection{Automatically Choosing a Model}

Based on our observations, we seek to automatically \linebreak choose a value of $\threshold$ that trades off between predictive performance and computational cost using the labeling functions' outputs $\lfout$ alone.
Including $\threshold$ as a hyperparameter in a grid search over a development set is generally not feasible because of its large effect on running time.
We therefore want to choose $\threshold$ before other hyperparameters, without performing any parameter estimation.
We propose using the number of correlations selected at each value of $\threshold$ as an inexpensive indicator.
The dashed lines in Figure~\ref{fig:sl_optimizer} show that as $\threshold$ decreases, the number of selected correlations follows a pattern.
Generally, the number of correlations grows slowly at first, then hits an ``elbow point'' beyond which the number explodes, which fits the assumption that the correlation structure is sparse.
In all three cases, setting $\threshold$ to this elbow point is a safe tradeoff between predictive performance and computational cost.
In cases where performance grows consistently (left and right), the elbow point achieves most of the predictive performance gains at a small fraction of the computational cost.
For example, on Spouses (right), choosing $\threshold = 0.08$ achieves a score of 56.6 F1---within one point of the best score---but only takes 8 minutes for parameter estimation.
In cases where predictive performance eventually degrades (middle), the elbow point also selects a relatively small number of correlations, giving an 0.7 F1 point improvement and avoiding overfitting.

Performing structure learning for many settings of $\threshold$ is inexpensive, especially since the search needs to be performed only once before tuning the other hyperparameters.
On the large number of labeling functions in the Spouses task, structure learning for 25 values of $\threshold$ takes 14 minutes.
On CDR, with a smaller number of labeling functions, it takes 30 seconds.
Further, if the search is started at a low value of $\threshold$ and increased, it can often be terminated early, when the number of selected correlations reaches a low value.
Selecting the elbow point itself is straightforward.
We use the point with greatest absolute difference from its neighbors, but more sophisticated schemes can also be applied \cite{satopaa:icdcs11}.
Our full optimization algorithm for choosing a modeling strategy and (if necessary) correlations is shown in Algorithm~\ref{alg:modeling-optimizer}.

\begin{algorithm}[tb]
\caption{Modeling Strategy Optimizer}
\label{alg:modeling-optimizer}
\begin{algorithmic}
   \vskip 0.2em
   \STATE {\bfseries Input:} 
   Label matrix $\lfout \in \left( \domy\cup\{\emptyset\} \right)^{\ny\times\nlf}$, \\
   advantage tolerance $\advtol$, structure search resolution $\searchres$
   \STATE {\bfseries Output:} Modeling strategy
   \vskip 0.4em
   \IF{$\tilde{A}^*(\lfout) < \advtol$} \RETURN {\bf MV}
   \ENDIF
     \STATE \texttt{Structures} $\leftarrow$ \texttt{[ ]}
     \FOR{$i$ {\bf from} 1 {\bf to} $\frac{1}{2 \searchres}$}
     \STATE $\threshold \leftarrow i \cdot \searchres$
     \STATE $C \leftarrow$ \texttt{LearnStructure(}$\lfout, \threshold$\texttt{)}
     \STATE \texttt{Structures.append(}$|C|, \threshold$\texttt{)}
     \ENDFOR
     \STATE $\threshold \leftarrow$ \texttt{SelectElbowPoint(Structures)}
   \RETURN {\bf GM}$_\threshold$
\end{algorithmic}
\end{algorithm}

%% file: experiments.tex

\section{Evaluation}
\label{sec:experiments}

We evaluate Snorkel by drawing on deployments developed in collaboration with users.
We report on two real-world deployments and four tasks on open-source data sets representative of other deployments.
Our evaluation is designed to support the following three main claims:
\begin{itemize}\compactify
    
    \item \textbf{Snorkel outperforms distant supervision baselines.}
    In \textit{distant supervision}~\cite{mintz:acl09}, one of the most popular forms of weak supervision used in practice, an external knowledge base is heuristically aligned with input data to serve as noisy training labels.
    By allowing users to easily incorporate a broader, more heterogeneous set of weak supervision sources, Snorkel exceeds models trained via distant supervision by an average of $\avgimprovementoverheuristic\%$.
    
    \item \textbf{Snorkel approaches hand supervision.}
    We see that by writing tens of labeling functions, we were able to approach or match results using hand-labeled training data which took weeks or months to assemble, coming within $\avgHLgapRE\%$ of the F1 score of hand supervision on relation extraction tasks and an average $\avgHLgapCM\%$ accuracy or AUC on cross-modal tasks, for an average $\avgHLgap\%$ across all tasks.

    \item \textbf{Snorkel enables  a new interaction paradigm.} We measure Snorkel's efficiency and ease-of-use by reporting on a user study of biomedical researchers from across the U.S.
    These participants learned to write labeling functions to extract relations from news articles as part of a two-day workshop on learning to use Snorkel, and matched or outperformed models trained on hand-labeled training data, showing the efficiency of Snorkel's process even for first-time users.
\end{itemize}

We now describe our results in detail.
First, we describe the six applications that validate our claims.
We then show that Snorkel's generative modeling stage helps to improve the predictive performance of the discriminative model, \linebreak demonstrating that it is $\avgrelimprovementoverMV\%$ more accurate when trained on Snorkel's probabilistic labels versus labels produced by an unweighted average of labeling functions.
We also validate that the ability to incorporate many different types of weak supervision incrementally improves results with an ablation study.
Finally, we describe the protocol and results of our user study.

\subsection{Applications}
\label{sec:applications}

\begin{table}[t]
\centering
\caption{Number of labeling functions, fraction of positive labels (for binary classification tasks), number of training documents, and number of training candidates for each task.}
\label{tab:dataset_stats}
\begin{tabular}{l r r r r}
\toprule
Task & \# LFs & \% Pos. & \# Docs & \# Candidates\\
\midrule
Chem & 16 & 4.1 & 1,753 & 65,398\\
EHR & 24 & 36.8 & 47,827 & 225,607 \\
CDR & 33 & 24.6 & 900 & 8,272 \\
Spouses & 11 & 8.3 & 2,073 & 22,195 \\
Radiology & 18 & 36.0 & 3,851 & 3,851 \\
Crowd & 102 & - & 505 & 505 \\
\bottomrule
\end{tabular}
\end{table}

To evaluate the effectiveness of Snorkel, we consider several real-world deployments and tasks on open-source datasets that are representative of other deployments in information extraction, medical image classification, and crowdsourced sentiment analysis.
Summary statistics of the tasks are provided in Table~\ref{tab:dataset_stats}.
\\ \\
\begin{table*}[t]
\centering
\caption{
    Evaluation of Snorkel on relation extraction tasks from text.
    Snorkel's generative and discriminative models consistently improve over distant supervision, measured in F1, the harmonic mean of precision (P) and recall (R).
    We compare with hand-labeled data when available, coming within an average of 1 F1 point.
}
\label{tab:benchmarks_text}
\begin{tabular}{l  ccc c  cccr  c cccr  c ccc}
\toprule
 & \multicolumn{3}{c}{Distant Supervision} & \phantom{a} & \multicolumn{4}{c}{Snorkel (Gen.)} & \phantom{a} & \multicolumn{4}{c}{Snorkel (Disc.)} & \phantom{a} & \multicolumn{3}{c}{Hand Supervision}\\
Task & P & R & {\bf F1} & & P & R & {\bf F1} & {\bf Lift} & & P & R & {\bf F1} & {\bf Lift} & & P & R & {\bf F1} \\
\midrule
Chem    & 11.2 & 41.2 & 17.6 & & 78.6 & 21.6 & 33.8 & +16.2 & & 87.0 & 39.2 & 54.1 & +36.5 & & - & - & - \\ 
EHR      & 81.4 & 64.8 & 72.2 & & 77.1 & 72.9 & 74.9  & +2.7 & & 80.2 & 82.6 & 81.4 & +9.2 & & - & - & - \\
CDR      & 25.5 & 34.8 & 29.4 & & 52.3 & 30.4 & 38.5 & +9.1 & & 38.8 & 54.3 & 45.3 & +15.9 &  & 39.9 & 58.1 & 47.3 \\  
Spouses & 9.9 & 34.8 & 15.4   & & 53.5 & 62.1 & 57.4 & +42.0 & & 48.4 & 61.6 & 54.2 & +38.8 & & 47.8 & 62.5 & 54.2\\
\bottomrule
\end{tabular}
\end{table*}

\noindent {\bf Discriminative Models:}
One of the key bets in Snorkel's design is that the trend of increasingly powerful, open-source machine learning tools (e.g., models, pre-trained word embeddings and initial layers, automatic tuners, etc.) will only continue to accelerate.
To best take advantage of this, Snorkel creates probabilistic training labels for any discriminative model with a standard loss function.

In the following experiments, we control for end model selection by using currently popular, standard choices across all settings.
For text modalities, we choose a bidirectional long short term memory (LSTM) sequence model~\cite{graves2005framewise}, and for the medical image classification task we use a 50-layer ResNet~\cite{DBLP:journals/corr/HeZRS15} pre-trained on the ImageNet object classification dataset~\cite{deng2009imagenet}.
Both models are implemented in Tensorflow \cite{abadi:osdi16} and trained using the Adam optimizer~\cite{kingma2014adam}, with hyperparameters selected via random grid search using a small labeled development set.
Final scores are reported on a held-out labeled test set.
See full version for details.

A key takeaway of the following results is that the discriminative model generalizes beyond the heuristics encoded in the labeling functions (as in Example~\ref{ex:generalization}).
In Section~\ref{sec:relation_extraction}, we see that on relation extraction applications the discriminative model improves performance over the generative \linebreak model primarily by increasing recall by $\avgrecallimprovementDM\%$ on average.
In Section~\ref{sec:benchmarks_cross_modal}, the discriminative model classifies entirely new modalities of data to which the labeling functions cannot be  applied.

\subsubsection{Relation Extraction from Text}
\label{sec:relation_extraction}

We first focus on four relation extraction tasks on text data, as it is a challenging and common class of problems that are well studied and for which distant supervision is often considered.
Predictive performance is summarized in Table~\ref{tab:benchmarks_text}.
We briefly describe each task.
\vspace{-.5em} \\
\\
\noindent {\bf Scientific Articles (Chem):}
With modern online repositories of scientific literature, such as PubMed\footnote{\small{\url{https://www.ncbi.nlm.nih.gov/pubmed/}}} for biomedical articles, research results are more accessible than ever before.
However, actually extracting fine-grained pieces of information in a structured format and using this data to answer specific questions at scale remains a significant open challenge for researchers.
To address this challenge in the context of drug safety research, Stanford and U.S. Food and Drug Administration (FDA) collaborators used Snorkel to develop a system for extracting chemical reagent and reaction product relations from PubMed abstracts.
\linebreak The goal was to build a database of chemical reactions that researchers at the FDA can use to predict unknown drug interactions.
We used the chemical reactions described in the Metacyc database \cite{metacyc} for distant supervision.
\\ \\
\noindent {\bf Electronic Health Records (EHR):}
As patients' clinical records increasingly become digitized, researchers hope to inform clinical decision making by retrospectively analyzing large patient cohorts, rather than conducting expensive randomized controlled studies.
However, much of the valuable information in electronic health records (EHRs)---such as fine-grained clinical details, practitioner notes, etc.---is not contained in standardized medical coding systems, and is thus locked away in the unstructured text notes sections.
In collaboration with researchers and clinicians at the U.S. Department of Veterans Affairs, Stanford Hospital and Clinics (SHC), and the Stanford Center for Biomedical Informatics Research, we used Snorkel to develop a system to extract structured data from unstructured EHR notes. 
Specifically, the system's task was to extract mentions of pain levels at precise anatomical locations from clinician notes, with the goal of using these features to automatically assess patient well-being and detect complications after medical interventions like surgery.
To this end, our collaborators created a cohort of 5,800 patients from SHC EHR data, with visit dates between 1995 and 2015, resulting in 500K unstructured clinical documents.
Since distant supervision from a knowledge base is not applicable, we compared against regular-expression-based labeling previously developed for this task.
\\ \\
\noindent {\bf Chemical-Disease Relations (CDR):}
We used the 2015 BioCreative chemical-disease relation dataset~\cite{wei:biocreative15}, where the task is to identify mentions of causal links between chemicals and diseases in PubMed abstracts.
We used all pairs of chemical and disease mentions co-occuring in a sentence as our candidate set.
We used the Comparative Toxicogenomics Database (CTD) \cite{davis:nar16} for distant supervision, and additionally wrote labeling functions  capturing language patterns and information from the context hierarchy.
To evaluate Snorkel's ability to discover previously unknown information, we randomly removed half of the relations in CTD and evaluated on candidates not contained in the remaining half.
\vspace{-.5em} \\ \\
\noindent {\bf Spouses:}
Our fourth task is to identify mentions of spouse relationships in a set of news articles from the Signal Media dataset \cite{corney:ecirws16}.
We used all pairs of person mentions (tagged with SpaCy's NER module\footnote{\small{\url{https://spacy.io/}}}) co-occuring in the same sentence as our candidate set.
To obtain hand-labeled data for evaluation, we crowdsourced labels for the candidates via Amazon Mechanical Turk, soliciting labels from three workers for each example and assigning the majority vote.
We then wrote labeling functions that encoded language patterns and distant supervision from DBpedia \cite{lehmann:semanticweb14}.

\subsubsection{Cross-Modal: Images \& Crowdsourcing}
\label{sec:benchmarks_cross_modal}

\begin{table}[t]
\centering
\caption{Evaluation on cross-modal experiments. Labeling functions that operate on or represent one modality (text, crowd workers) produce training labels for models that operate on another modality (images, text), and approach the predictive performance of large hand-labeled training datasets.}
\label{tab:benchmarks_xmodal}
\begin{tabular}{l c c}
\toprule
Task & Snorkel (Disc.) & Hand Supervision \\
\midrule
Radiology (AUC) &  72.0 & 76.2 \\
Crowd (Acc) & 65.6 & 68.8\\
\bottomrule
\end{tabular}
\end{table}

In the cross-modal setting, we write labeling functions over one data modality (e.g., a text report, or the votes of crowdworkers) and use the resulting labels to train a classifier defined over a second, totally separate modality (e.g., an image or the text of a tweet).
This demonstrates the flexibility of Snorkel, in that the labeling functions (and by extension, the generative model) do not need to operate over the same domain as the discriminative model being trained.
Predictive performance is summarized in Table~\ref{tab:benchmarks_xmodal}.
\\ \\
\noindent {\bf Abnormality Detection in Lung Radiographs (Rad):}
In many real-world radiology settings, there are large repositories of image data with corresponding narrative text reports, but limited or no labels that could be used for training an image classification model.
In this application, in collaboration with radiologists, we wrote labeling functions over the text radiology reports, and used the resulting labels to train an image classifier to detect abnormalities in lung X-ray images.
We used a publicly available dataset from the OpenI biomedical image repository\footnote{\small{\url{http://openi.nlm.nih.gov/}}} consisting of 3,851 distinct radiology reports---composed of unstructured text and Medical Subject Headings (MeSH)\footnote{\small{\url{https://www.nlm.nih.gov/mesh/meshhome.html}}} codes---and accompanying X-ray images.
\\ \\ \\
\noindent {\bf Crowdsourcing (Crowd):}
We trained a model to perform sentiment analysis using crowdsourced annotations from the weather sentiment task from Crowdflower.\footnote{\small{\url{https://www.crowdflower.com/data/weather-sentiment/}}}
In this task, contributors were asked to grade the sentiment of often-ambiguous tweets relating to the weather, choosing between five categories of sentiment.
Twenty contributors graded each tweet, but due to the difficulty of the task and lack of crowdworker filtering, there were many conflicts in worker labels.
We represented each crowdworker as a labeling \linebreak function---showing Snorkel's ability to subsume existing \linebreak crowdsourcing modeling approaches---and then used the resulting labels to train a text model over the tweets, for making predictions independent of the crowd workers.

\subsubsection{Effect of Generative Modeling}

An important question is the significance of modeling the accuracies and correlations of the labeling functions on the end predictive performance of the discriminative model (versus in Section~\ref{sec:modeling}, where we only considered the effect on the accuracy of the generative model).
We compare Snorkel with a simpler pipeline that skips the generative modeling stage and trains the discriminative model on an unweighted average of the labeling functions' outputs.
Table~\ref{tab:dmv_vs_dgm} shows that the discriminative model trained on Snorkel's probabilistic labels consistently predicts better, improving $\avgrelimprovementoverMV\%$ on average.
These results demonstrate that the discriminative model effectively learns from the additional signal contained in Snorkel's probabilistic training labels over simpler modeling strategies.

\begin{table}[t]
\centering
\caption{Comparison between training the discriminative model on the labels estimated by the generative model, versus training on the unweighted average of the LF outputs. Predictive performance gains show that modeling LF noise helps.}
\label{tab:dmv_vs_dgm}
\begin{tabular}{l c c r}
\toprule
& Disc. Model on & & \\
Task & Unweighted LFs & Disc. Model & Lift\\
\midrule
Chem & 48.6 & 54.1 & +5.5 \\
EHR & 80.9 & 81.4 & +0.5 \\
CDR & 42.0 & 45.3 & +3.3 \\
Spouses & 52.8 & 54.2 & +1.4 \\
Crowd (Acc) & 62.5 & 65.6 & +3.1 \\
Rad. (AUC) & 67.0 & 72.0 & +5.0 \\
\bottomrule
\end{tabular}
\end{table}

\subsubsection{Labeling Function Type Ablation}

We also examine the impact of different types of labeling functions on end predictive performance, using the CDR application as a representative example of three common categories of labeling functions:
\begin{itemize}\compactify
    \item \textit{Text Patterns:} Basic word, phrase, and regular expression labeling functions.
    
    \item \textit{Distant Supervision:} External knowledge bases mapped to candidates, either directly or filtered by a heuristic.
    
    \item \textit{Structure-Based:} Labeling functions expressing heuristics over the context hierarchy, e.g., reasoning about position in the document or relative to other candidates.
\end{itemize}
We show an ablation in Table~\ref{tab:lf_ablation}, sorting by stand-alone score.
We see that distant supervision adds recall at the cost of some precision, as we would expect, but ultimately improves F1 score by 2 points; and that structure-based labeling functions, enabled by Snorkel's context hierarchy data representation, add an additional F1 point.

\begin{table}[t]
\centering
\caption{Labeling function ablation study on CDR. Adding different types of labeling functions improves predictive performance.}
\label{tab:lf_ablation}
\begin{tabular}{l c c c r}
\toprule
LF Type & P & R & F1 & Lift\\
\midrule
Text Patterns & 42.3 & 42.4 & 42.3 & \\
+ Distant Supervision & 37.5 & 54.1 & 44.3 & +2.0\\
+ Structure-based & 38.8 & 54.3 & 45.3 & +1.0\\
\bottomrule
\end{tabular}
\end{table}

\subsection{User Study}
\label{sec:user_study}

We conducted a formal study of Snorkel to (i) evaluate how quickly SME users could learn to write labeling functions, and (ii) empirically validate the core hypothesis that writing labeling functions is more time-efficient than hand-labeling data.
Users were given instruction on Snorkel, and then asked to write labeling functions for the Spouses task described in the previous subsection.
\\ \\
\noindent {\bf Participants:}
In collaboration with the Mobilize Center \cite{ku2015mobilize}, an NIH-funded Big Data to Knowledge (BD2K) center, we distributed a national call for applications to attend a two-day workshop on using Snorkel for biomedical knowledge base construction. 
Selection criteria included a strong biomedical project proposal and little-to-no prior experience using Snorkel. 
In total, 15 researchers\footnote{\small{One participant declined to write labeling functions, so their score is not included in our analysis.}} were invited to attend out of 33 team applications submitted, with varying backgrounds in bioinformatics, clinical informatics, and data mining from universities, companies, and organizations around the United States. 
The education demographics included 6 bachelors, 4 masters, and 5 Ph.D. degrees. All participants could program in Python, with 80\% rating their skill as intermediate or better; 40\% of participants had little-to-no prior exposure to machine learning; and 53-60\% had no prior experience with text mining or information extraction applications.
See full version for details.
\\ \\
\noindent {\bf Protocol:}
The first day focused entirely on labeling functions, ranging from theoretical motivations to details of the Snorkel API.
Over the course of 7 hours, participants were instructed in a classroom setting on how to use and evaluate models developed using Snorkel.
Users were presented with 4 tutorial Jupyter notebooks providing skeleton code for evaluating labeling functions, along with a small labeled development candidate set, and were given 2.5 hours of dedicated development time in aggregate to write their labeling functions.
All workshop materials are available online.\footnote{\small{\url{https://github.com/HazyResearch/snorkel/tree/master/tutorials/workshop}}}
\\ \\
\noindent {\bf Baseline:}
To compare our users' performance against models trained on hand-labeled data, we collected a large hand-labeled dataset via Amazon Mechanical Turk (the same set used in the previous subsection).
We then split this into 15 datasets representing 7 hours worth of hand-labeling time each---based on the crowd-worker average of 10 seconds per label---simulating the alternative scenario where users \linebreak skipped both instruction and labeling function development sessions and instead spent the full day hand-labeling data.
\\ \\
\noindent {\bf Results:}
Our key finding is that labeling functions written in Snorkel, even by SME users, can match or exceed a traditional hand-labeling approach.
The majority (8) of subjects matched or outperformed these hand-labeled data models.
The average Snorkel user's score was 30.4 F1, and the average hand-supervision score was 20.9 F1.
The best performing user model scored 48.7 F1, 19.2 points higher than the best supervised model using hand-labeled data. The worst participant scored 12.0 F1, 0.3 points higher that the lowest hand-labeled model.
The full distribution of scores by participant, and broken down by participant background, compared against the baseline models trained with hand-labeled data is available in the full version.

%% file: related.tex
\vspace{-.5em}
\section{Related Work}
\label{sec:related}

This section is an overview of techniques for managing weak supervision, many of which are subsumed in Snorkel.
We also contrast it with related forms of supervision.
\\ \\ 
\noindent{\bf Combining Weak Supervision Sources:}
The main challenge of weak supervision is how to combine multiple sources.
For example, if a user provides two knowledge bases for distant supervision, how should a data point that matches only one knowledge base be labeled?
Some researchers have used multi-instance learning to reduce the noise in weak supervision sources \cite{riedel:ecml10, hoffmann:acl11}, essentially modeling the different weak supervision sources as soft constraints on the true label, but this approach is limited because it requires using a specific end model that supports multi-instance learning.

Researchers have therefore considered how to estimate the accuracy of label sources without a gold standard with which to compare---a classic problem \cite{dawid:royalstats79}---and combine these estimates into labels that can be used to train an arbitrary end model.
Much of this work has focused on crowdsourcing, in which workers have unknown accuracy \cite{dalvi:www13, joglekar:icde15, zhang:jmlr16}.
Such methods use generative probabilistic models to estimate a latent variable---the true class label---based on noisy observations.
Other methods use generative models with hand-specified dependency structures to label data for specific modalities, such as topic models for text \cite{alfonseca:acl12} or denoising distant supervision sources \cite{takamatsu:acl12, roth:emnlp13}.
Other techniques for estimating latent class labels given noisy observations include spectral methods \cite{parisi:pnas14}.
Snorkel is distinguished from these approaches because its generative model supports a wide range of weak supervision sources, and it learns the accuracies and correlation structure among weak supervision sources without ground truth data.
\\ \\
\noindent {\bf Other Forms of Supervision:}
Work on \textit{semi-supervised learning} considers settings with some labeled data and a much larger set of unlabeled data, and then leverages various domain- and task-agnostic assumptions about smoothness, low-dimensional structure, or distance metrics to heuristically label the unlabeled data~\cite{chapelle:09}.
Work on \textit{active learning} aims to automatically estimate which data points are optimal to label, thereby hopefully reducing the total number of examples that need to be manually annotated~\cite{settles:12}.
\textit{Transfer learning} considers the strategy of repurposing models trained on different datasets or tasks where labeled training data is more abundant~\cite{pan:tkde10}.
Another type of supervision is self-training \cite{scudder:infotheory65, agrawala:infotheory70} and co-training \cite{blum:colt98}, which involves training a model or pair of models on data that they labeled themselves.
Weak supervision is distinct in that the goal is to solicit input directly from SMEs, however at a higher level of abstraction and/or in an inherently noisier form.
Snorkel is focused on managing weak supervision sources, but combing its methods with these other types of supervision is straightforward.
\\ \\
\noindent {\bf Related Data Management Problems:}
Researchers \linebreak have considered related problems in data management, such as data fusion \cite{dong:book15, rekatsinas:sigmod17} and truth discovery \cite{li:kddex15}.
In these settings, the task is to estimate the reliability of data sources that provide assertions of facts and determine which facts are likely true.
Many approaches to these problems use probablistic graphical models that are related to Snorkel's generative model in that they represent the unobserved truth as a latent variable, e.g., the latent truth model \cite{zhao:vldb12}.
Our setting differs in that labeling functions assign labels to user-provided data, and they may provide any label or abstain, which we must model.
Work on data fusion has also explored how to model user-specified correlations among data sources \cite{pochampally:sigmod14}.
Snorkel automatically identifies which correlations among labeling functions to model.

%% file: conclusion.tex

\section{Conclusion}
\label{sec:conclusion}

Snorkel provides a new paradigm for soliciting and managing weak supervision to create training data sets.
In Snorkel, users provide higher-level supervision in the form of labeling functions that capture domain knowledge and resources, without having to carefully manage the noise and conflicts inherent in combining weak supervision sources.
Our evaluations demonstrate that Snorkel significantly reduces the cost and difficulty of training powerful machine learning models while exceeding prior weak supervision methods and approaching the quality of large, hand-labeled training sets.
Snorkel's deployments in industry, research labs, and government agencies show that it has real-world impact, offering developers an improved way to build models.

%% file: appendix.tex

\pagebreak

\begin{appendix}

\section{Additional Material for Sec. 3.1}
\subsection{Minor Notes}
Note  that for the independent generative model (i.e., $|C|=0$), the weight corresponding to the accuracy factor, $\genparam_\ilf$, for labeling function $\ilf$ is just the log-odds of its accuracy:
\begin{align*}
\alpha_\ilf &= P(\lfout_{\iy,\ilf}=1~|~Y_\iy=1, \lfout_{\iy,\ilf} \neq 0)\\
&= \frac{
  P(\lfout_{\iy,\ilf}=1, Y_\iy=1, \lfout_{\iy,\ilf} \neq 0)
  }{
  P(Y_\iy=1, \lfout_{\iy,\ilf} \neq 0)
  }\\
&= \frac{
  \exp(\genparam_\ilf)
  }{
  \exp(\genparam_\ilf) + \exp(-\genparam_\ilf)
  }\\
\implies \genparam_\ilf &= \frac12\log\left( 
    \frac{\alpha_\ilf}{1-\alpha_\ilf} 
  \right)
\end{align*}
Also note that the accuracy we consider is conditioned on the labeling function not abstaining, i.e.,:
\begin{align*}
P(\lfout_{\iy,\ilf}=1~|~Y_\iy=1) = \alpha_\ilf * P(\lfout_{\iy,\ilf}\neq\emptyset)
\end{align*}
because a separate factor $\dep_{\iy, \ilf}^{\textrm{Lab}}$ captures how often each labeling function votes.

\subsection{Proof of Proposition 1}
In this proposition, our goal is to obtain a simple upper bound for the expected optimal advantage $\mathbb{E}_{\lfout,y,\genparam^*}\left[ A^* \right]$ in the low label density regime.
We consider a simple model where all the labeling functions have a fixed probability of emitting a non-zero label, 
\begin{align}
P(\lfout_{\iy,\ilf}\neq\emptyset) = p_l\ \forall \iy,\ilf
\end{align}
and that the labeling functions are all non-adversarial, i.e., they all have accuracies greater than 50\%, or equivalently,
\begin{align}
\genparam^*_\ilf > 0\ \forall \ilf \label{assumption:non-adv}
\end{align}
First, we start by only counting cases where the optimal weighted majority vote (WMV*)---i.e., the predictions of the generative model with perfectly estimated weights---is correct and the majority vote (MV) is incorrect, which is an upper bound on the modeling advantage:
\begin{align*}
&\mathbb{E}_{\lfout,y,\genparam^*}\left[ A_{\genparam^*}(\lfout, y) \right]\\
&= 
  \frac{1}{\ny}\sum_{\iy=1}^\ny\left(
      \mathbb{E}_{\lfout_i,y_i,\genparam^*}\left[
        \mathbbm{1}\left\{
            y_\iy f_{\genparam^*}(\lfout_\iy) > 0 \wedge y_\iy f_1(\lfout_\iy) \leq 0
        \right\}\right.\right.\\
       -&\left.\left.\mathbbm{1}\left\{y_\iy f_{\genparam^*}(\lfout_\iy) \leq 0 \wedge y_\iy f_1(\lfout_\iy) > 0 \right\}
       \right]\right)\\
&\leq 
  \frac{1}{\ny}\sum_{\iy=1}^\ny\mathbb{E}_{\lfout_i,y_i,\genparam^*}\left[
        \mathbbm{1}\left\{
            y_\iy f_{\genparam^*}(\lfout_\iy) > 0 \wedge y_\iy f_1(\lfout_\iy) \leq 0
        \right\}\right]
\end{align*}
Next, by (\ref{assumption:non-adv}), the only way that WMV* and MV could possibly disagree is if there is at least one disagreeing pair of labels:
\begin{align*}
\mathbb{E}_{\lfout,y,\genparam^*}\left[ A^*(\lfout, y) \right]
&\leq \frac{1}{\ny}\sum_{\iy=1}^\ny\mathbb{E}_{\lfout,y}\left[ 
        \mathbbm{1}\left\{
            c_1(\lfout_i) > 0 \wedge c_{-1}(\lfout_i) > 0
        \right\}\right]
\end{align*}
where $c_y(\lfout_\iy) = \sum_{\ilf=1}^\nlf \mathbbm{1}\left\{ \lfout_{\iy,\ilf} = y\right\}$, in other words, the counts of positive or negative labels for a given data point $x_\iy$.
Then, we can bound this by the expected number of disagreeing, non-abstaining pairs of labels:
\begin{align*}
\mathbb{E}&_{\lfout,y,\genparam^*}\left[ A^*(\lfout, y) \right]\\
&\leq \frac{1}{\ny}\sum_{\iy=1}^\ny\mathbb{E}_{\lfout,y}\left[ 
        \sum_{\ilf=1}^{\nlf-1} \sum_{\ilfalt=\ilf+1}^\nlf
        \mathbbm{1}\left\{
            \lfout_{\iy,\ilf}\neq\lfout_{\iy,\ilfalt} \wedge \lfout_{\iy,\ilf},\lfout_{\iy,\ilfalt}\neq0
        \right\}\right]\\
&= \frac{1}{\ny}\sum_{\iy=1}^\ny 
        \sum_{\ilf=1}^{\nlf-1} \sum_{\ilfalt=\ilf+1}^\nlf
        \mathbb{E}_{\lfout,y}\left[
        \mathbbm{1}\left\{
            \lfout_{\iy,\ilf}\neq\lfout_{\iy,\ilfalt} \wedge \lfout_{\iy,\ilf},\lfout_{\iy,\ilfalt}\neq0
        \right\}\right]\\
&= \frac{1}{\ny}\sum_{\iy=1}^\ny 
        \sum_{\ilf=1}^{\nlf-1} \sum_{\ilfalt=\ilf+1}^\nlf
        \sum_{y'\in\pm 1}\sum_{\lambda\in\pm 1}
        P(\lfout_{\iy,\ilf}=\lambda,\lfout_{\iy,\ilfalt}=-\lambda, y_i=y')
\end{align*}
Since we are considering the independent model, $\lfout_{\iy,\ilf}\perp\lfout_{\iy,\ilfalt\neq\ilf}\ |\ y_\iy$, we have that:
\begin{align*}
&P(\lfout_{\iy,\ilf}=\lambda,\lfout_{\iy,\ilfalt}=-\lambda, y_i=\lambda)\\
&= P(\lfout_{\iy,\ilf}=\lambda~|~y_i=\lambda) 
    P(\lfout_{\iy,\ilfalt}=-\lambda~|~y_i=\lambda) P(y_i=\lambda)\\
&= \alpha_\ilf(1-\alpha_\ilfalt) p_l^2 P(y_i=\lambda)
\end{align*}
Thus we have:
\begin{align*}
\mathbb{E}_{\lfout,y,\genparam^*}\left[ A^*(\lfout, y) \right]
&\leq \sum_{\ilf=1}^{\nlf-1} \sum_{\ilfalt=\ilf+1}^\nlf
        p_l^2\left(
            \alpha_\ilf(1-\alpha_\ilfalt)
            + (1-\alpha_\ilf)\alpha_\ilfalt
        \right)\\
&= \sum_{\ilf=1}^{\nlf} \sum_{\ilfalt \neq \ilf}
        p_l^2 \alpha_\ilf(1-\alpha_\ilfalt)\\
&\leq \sum_{\ilf=1}^{\nlf} \sum_{\ilfalt=1}^{\nlf}
        p_l^2 \alpha_\ilf(1-\alpha_\ilfalt)\\
&= n^2p_l^2 \bar{\alpha}(1-\bar{\alpha})\\
&= \bar{d}^2\bar{\alpha}(1-\bar{\alpha})
\end{align*}
where we have defined the average labeling function accuracy as $\bar{\alpha}$, and where the label density is defined as $\bar{d}=p_l\nlf$.
Thus we have shown that the expected advantage scales at most quadratically in the label density. \hfill $\square$

\subsection{Explanation of Theorem 1}
The Dawid-Skene model~\cite{dawid:royalstats79} of crowd workers classically models each crowd worker as conditionally independent, and having some class-dependent but data point-independent probability of emitting a correct label.
In our setting, considering the binary classification case (as Dawid-Skene treats), we can think of each crowd worker as a labeling function, in which case we have:
\begin{align*}
\alpha^+_\ilf &= P(\lfout_{\iy,\ilf}=1\ |\ y_\iy=1, \lfout_{\iy,\ilf} \neq 0)\\
\alpha^-_\ilf &= P(\lfout_{\iy,\ilf}=-1\ |\ y_\iy=-1, \lfout_{\iy,\ilf} \neq 0)
\end{align*}
The \textit{symmetric} Dawid-Skene setting is the one we consider, where $\alpha_\ilf \equiv \alpha_\ilf^+ = \alpha_\ilf^-$.
Furthermore, we can refer to the matrix of probabilities of a worker $\ilf$ being given data point $\iy$ to label (i.e., in our syntax, the probability of not abstaining) as the \textit{sampling probability matrix}.
If the entries are all the same, this is refered to as a \textit{constant probability sampling strategy}, and is equivalent to our assumption $P(\lfout_{\iy,\ilf}\neq\emptyset) = p_l\ \forall \iy,\ilf$.

In this setting, and assuming that the mean labeling function / crowd worker accuracy $\bar{\alpha} > \frac12$ (or equivalently, $\bar{\genparam}^*>0$), then Corrollary 9 in~\cite{li2013error} provides us with the following upper bound on the mean error rate:
\begin{align}
\frac{1}{\ny} P(f_1(\lfout_{\iy})\neq y_\iy) \leq e^{-2\nlf p_l^2 (\bar{\alpha}^*-\frac12)^2} \label{eqn:corrli9}
\end{align}

We then use~(\ref{eqn:corrli9}) to upper bound the expected advantage \linebreak $\mathbb{E}_{\lfout,y,\genparam^*}\left[ A^* \right]$, substituting in $\bar{d}=\nlf p_l$ for clarity.

\subsection{Proof of Proposition 2}
In this proposition, our goal is to find a tractable upper bound on the conditional modeling advantage, i.e., the modeling advantage given the observed label matrix $\lfout$.
This will be useful because, given our label matrix, we can compute this quantity and, when it is small, safely skip learning the generative model and just use an unweighted majority vote (MV) of the labeling functions.
We assume in this proposition that the true weights of the labeling functions lie within a fixed range, $\genparam_\ilf\in [\genparam_{min}>0,\genparam_{max}]$ and have a mean $\bar{\genparam}$.
For notational convenience, let
\begin{align*}
y' = \begin{cases}
    1 & f_1(\lfout) > 0\\
    0 & f_1(\lfout) = 0\\
    -1 & f_1(\lfout) < 0\\
  \end{cases}
\end{align*}
We start with the expected advantage, and upper-bound by the expected number of instances in which WMV* is correct and MV is incorrect (note that for tie votes, we simply upper bound by trivially assuming an expected advantage of one):
\begin{align*}
\mathbb{E}&_{\genparam^*, y}
  \left[
    A^*(\lfout, y)~|~\lfout
  \right]\\
&=
\mathbb{E}_{\genparam^*, y \sim P(\cdot~|~\lfout, w^*)}
  \left[
    A_{\genparam^*}(\lfout, y)
  \right]\\
&\leq
\frac{1}{\ny} \sum_{\iy=1}^\ny 
  \mathbb{E}_{\genparam^*, y \sim P(\cdot~|~\lfout_\iy, \genparam^*)}
    \left[
      \mathbbm{1}\left\{ y_\iy \neq y'_\iy \right\}
      \mathbbm{1}\left\{ y'_\iy f_{\genparam^*}(\lfout_\iy) \leq 0 \right\}
    \right]\\
&=
\frac{1}{\ny} \sum_{\iy=1}^\ny 
  \mathbb{E}_{\genparam^*}
    \left[
      \mathbb{E}_{y \sim P(\cdot~|~\lfout_\iy, \genparam^*)}
        \left[
          \mathbbm{1}\left\{ y_\iy \neq y'_\iy \right\}
        \right]
      \mathbbm{1}\left\{ y'_\iy f_{\genparam^*}(\lfout_\iy) \leq 0 \right\}
    \right]\\
&=
\frac{1}{\ny} \sum_{\iy=1}^\ny 
  \mathbb{E}_{\genparam^*}
    \left[
      P(y_\iy \neq y'_\iy~|~\lfout_\iy, \genparam^*)
      \mathbbm{1}\left\{ y'_\iy f_{\genparam^*}(\lfout_\iy) \leq 0 \right\}
    \right]
\end{align*}
Next, define:
\begin{align*}
\Phi(\Lambda_\iy, y'')
&=
\mathbbm{1}
  \left\{  
    c_{y''}(\Lambda_\iy)\genparam_{max} - c_{-y''}(\Lambda_\iy)\genparam_{min}
  \right\}
\end{align*}
i.e. this is an indicator for whether WMV* could \textit{possibly} output $y''$ as a prediction under best-case circumstances.
We use this in turn to upper-bound the expected modeling advantage again:
\begin{align*}
\mathbb{E}&_{\genparam^*, y \sim P(\cdot~|~\lfout, w^*)}
  \left[
    A_{\genparam^*}(\lfout, y)
  \right]\\
&\leq
\frac{1}{\ny} \sum_{\iy=1}^\ny 
  \mathbb{E}_{\genparam^*}
    \left[
      P(y_\iy \neq y'_\iy~|~\lfout_\iy, \genparam^*)
      \Phi(\lfout_\iy, -y'_\iy)
    \right]\\
&=
\frac{1}{\ny} \sum_{\iy=1}^\ny
  \Phi(\lfout_\iy, -y'_\iy)
  \mathbb{E}_{\genparam^*}
    \left[
      P(y_\iy \neq y'_\iy~|~\lfout_\iy, \genparam^*)
    \right]\\
&\leq
\frac{1}{\ny} \sum_{\iy=1}^\ny
  \Phi(\lfout_\iy, -y'_\iy)
  P(y_\iy \neq y'_\iy~|~\lfout_\iy, \bar{\genparam})
\end{align*}
Now, recall that, for $y'\in\pm1$:
\begin{align*}
P(y_\iy=y'~|~\lfout_\iy, \genparam)
&= \frac{
    P(y_\iy=y', \lfout_\iy~|~\genparam)
  }{
    \sum_{y''\in\pm1}P(y_\iy=y'', \lfout_\iy~|~\genparam)
  }\\
&= \frac{
      \exp\left( \genparam^T \dep_\iy(\lfout_\iy, \y_\iy=y') \right)
    }{
      \sum_{y''\in\pm1}\exp\left( \genparam^T \dep_\iy(\lfout_\iy, \y_\iy=y'') \right)
    }\\
&= \frac{
      \exp\left( \genparam^T\lfout_\iy y' \right)
    }{
      \exp\left( \genparam^T\lfout_\iy \right) + \exp\left( -\genparam^T\lfout_\iy \right)
    }\\
&= \sigma\left( 2f_{\genparam}(\lfout_\iy)y' \right)
\end{align*}
where $\sigma(\cdot)$ is the sigmoid function.
Note that we are considering a simplified independent generative model with only accuracy factors;
however, in this discriminative formulation the labeling propensity factors would drop out anyway since they do not depend on $y$, so their omission is just for notational simplicity.

Putting this all together by removing the $y'_\iy$ placeholder, simplifying notation to match the main body of the paper, we have:
\begin{align*}
\mathbb{E}&_{\genparam^*, y}
  \left[
    A^*(\lfout, y)~|~\lfout
  \right]\\
&\leq
\frac{1}{\ny} \sum_{\iy=1}^\ny \sum_{y \in \pm1}
  \mathbbm{1}\left\{ y f_1(\lfout_\iy) \leq 0 \right\}
  \Phi(\lfout_\iy, y)
  \sigma
    \left( 
      2 y f_{\bar{\genparam}}(\lfout_\iy)
    \right)\\
&= \tilde{A}^*(\lfout)~\square.
\end{align*}

\subsection{Modeling Advantage Notes}

\begin{figure}[t]
\centering
\includegraphics[width=0.5\textwidth]{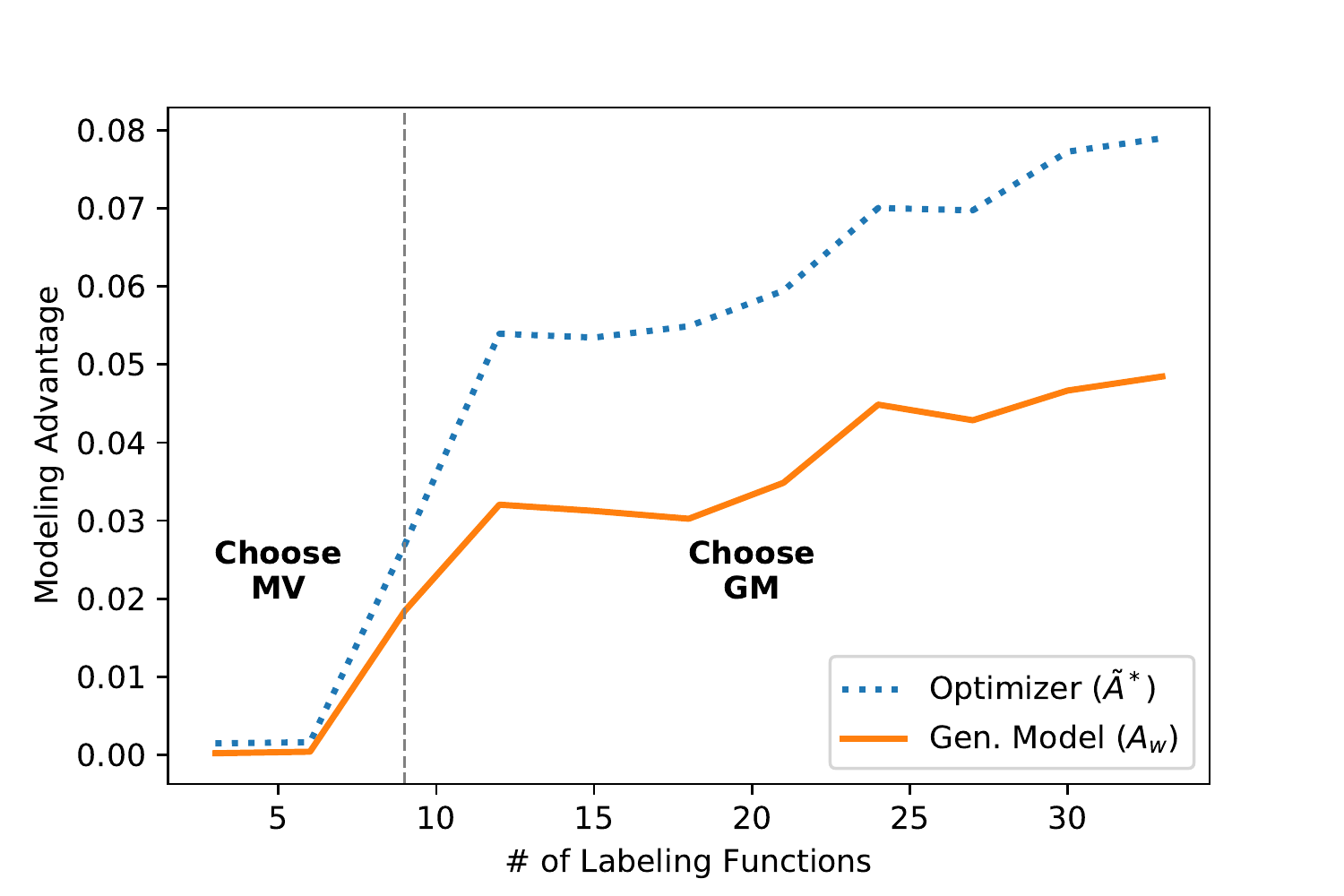}
\vspace{-20pt}
\caption[...]{
The advantage of using the generative labeling model (GM) over majority vote (MV) as predicted by our optimizer ($\tilde{A}^*$), and empirically ($A_w$), on the CDR application as the number of LFs is increased.
We see that the optimizer correctly chooses MV during early development stages, and then GM in later ones.
}
\vspace{-10pt}
\label{fig:label-density-regimes}
\end{figure}

In Figure~\ref{fig:label-density-regimes}, we measure the modeling advantage of the generative model versus a majority vote of the labeling functions on random subsets of the CDR labeling functions of different sizes.
We see that the modeling advantage grows as the number of labeling functions increases, indicating that the optimizer can save execution time especially during the initial stages of iterative development.

Note that in Section~\ref{sec:experiments}, due to known negative class imbalance in relation extraction problems, we count instances in which the generative model emits no label---i.e., a 0 label---as negatives, as is common practice (essentially, we are giving the generative model the benefit of the doubt given the known class imbalance).
Thus our reported F1 score metric hides instances in which the generative model learns to apply a -1 label where majority vote applied 0.
In computing the empirical modeling advantage, however, we \textit{do} count such instances as improvements over majority vote, as these instances \textit{do} have an effect on the training of the end discriminative model.

\section{Additional Evaluation Details}

\subsection{Data Set Details}

Additional information about the sizes of the datasets are included in Table~\ref{tab:additional_sizes}.
Specifically, we report the size of the (unlabeled) training set and hand-labeled development and test sets, in terms of number of candidates.
Note that the development and test sets can be orders of magnitude smaller that the training sets.
Labeled development and test sets were either used when already available as part of a benchmark dataset, or labeled with the help of our SME collaborators, limited to several hours of labeling time maximum.

\begin{table}[t]
\centering
\caption{Number of candidates in the training, development, and test splits for each dataset.}
\label{tab:additional_sizes}
\begin{tabular}{l r r r}
\toprule
Task & \# Train. & \# Dev. & \# Test \\
\midrule
Chem & 65,398 & 1,292 & 1,232 \\
EHR & 225,607 & 913 & 604 \\
CDR & 8,272 & 888 & 4,620 \\
Spouses & 22,195 & 2,796 & 2,697 \\
Radiology & 3,851 & 385 & 385 \\
Crowd & 505 & 63 & 64\\
\bottomrule
\end{tabular}
\end{table}

\subsection{User Study}
 
\begin{table}[t!]
\centering
\caption{Self-reported skill levels---beginner (Beg.), intermediate (Int.), and advanced (Adv.)---for all user study participants.}
\label{tab:userstudy}
\begin{tabular}{ccccc}
\toprule
Subject & New & Beg. & Int. & Adv. \\
\midrule
Python                  &  0  & 3 & 8 & 4 \\
Machine Learning        &  5  & 1 & 4 & 5 \\
Info. Extraction  &  2  & 6 & 5 & 2 \\
Text Mining             &  3  & 6 & 4 & 2 \\
\bottomrule
\end{tabular}
\end{table}

\begin{figure}[t!]
\centering
\includegraphics[width=0.45\textwidth,clip]{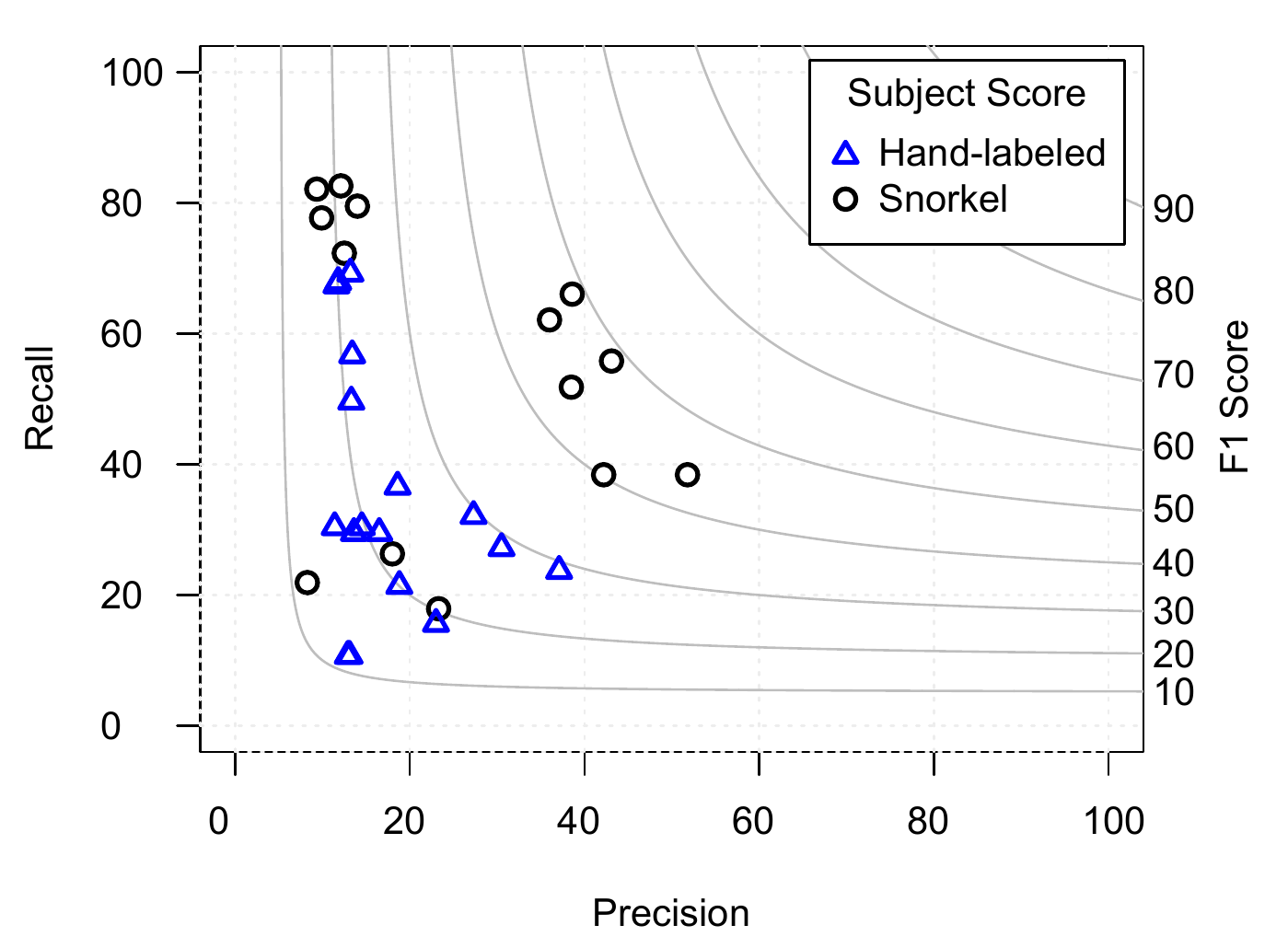}
\caption{Predictive performance of our 14 user study participants. The majority of users matched or exceeded the performance of a model trained on 7 hours (2500 instances) of hand-labeled data. }
\label{fig:userstudy}
\end{figure}

\begin{figure}[t!]
\centering
\includegraphics[width=0.2\textwidth,clip]{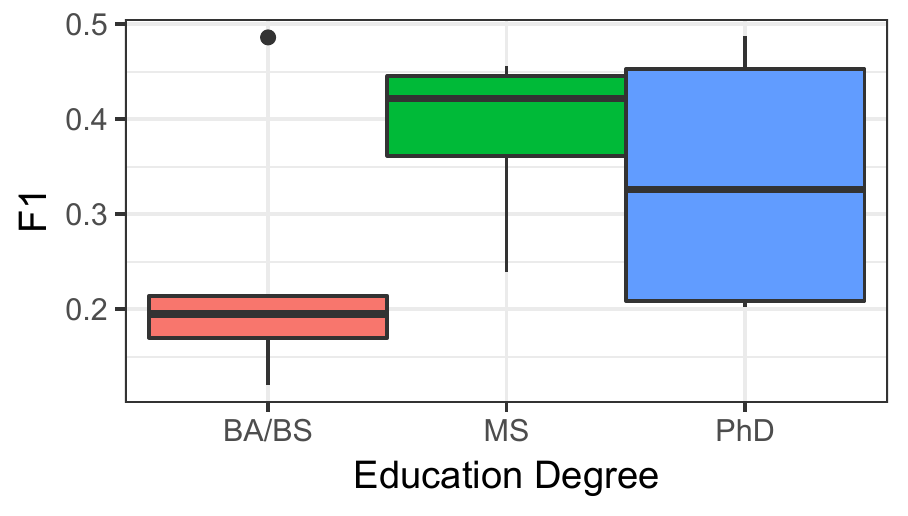}
\includegraphics[width=0.2\textwidth,clip]{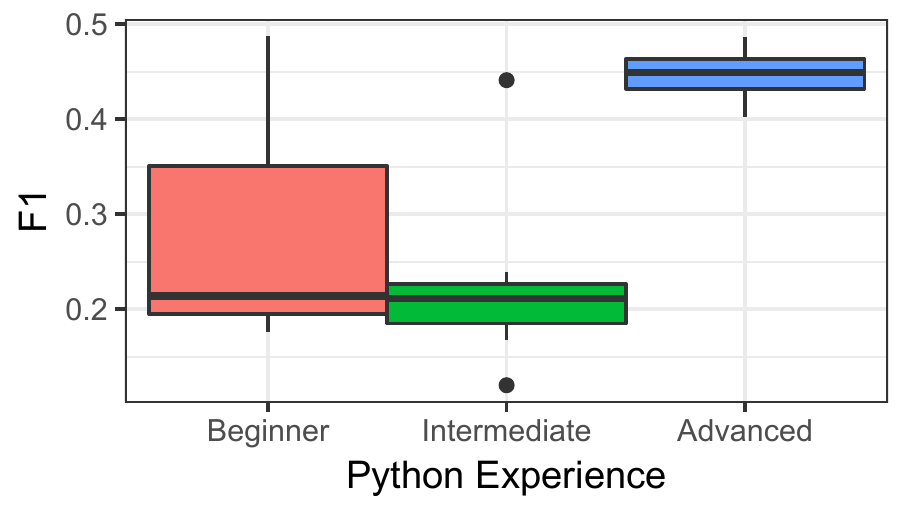}
\includegraphics[width=0.2\textwidth,clip]{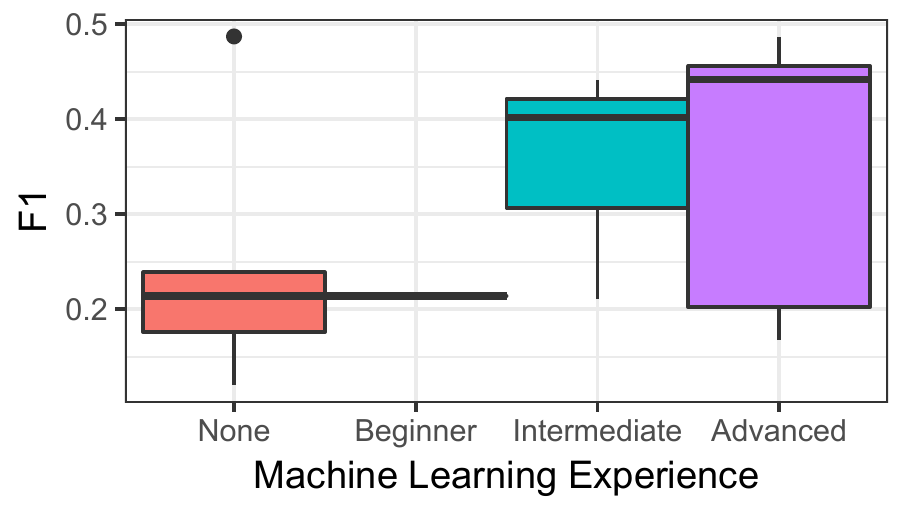}
\includegraphics[width=0.2\textwidth,clip]{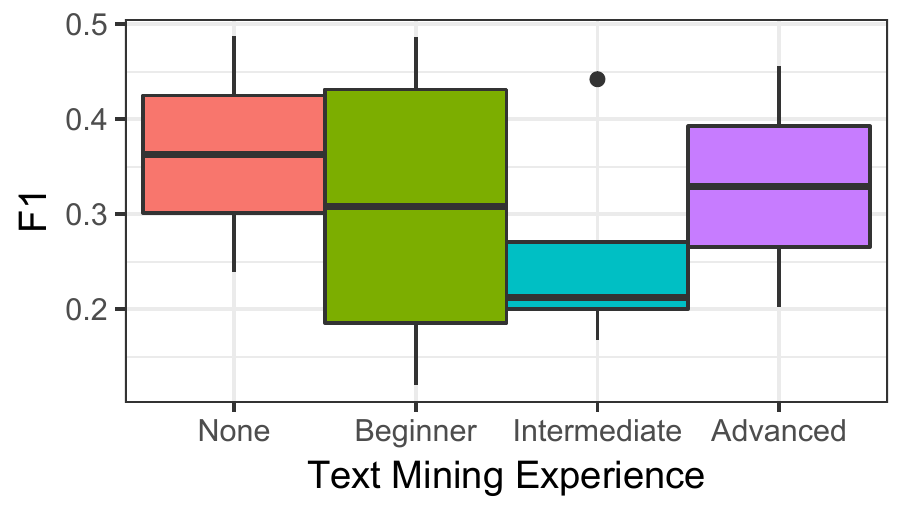}
\caption{The profile of the best performing user by F1 score, was a MS or Ph.D. degree in any field, strong Python coding skills, and intermediate to advanced experience with machine learning. Prior experience with text mining added no benefit.}
\label{fig:userstudy-factors}
\end{figure}

Figures \ref{fig:userstudy} and \ref{fig:userstudy-factors} show the distribution of scores by participant, and broken down by participant background, compared against the baseline models trained with hand-labeled data.
Figure \ref{fig:userstudy-factors} shows descriptive statistics of user factors broken down by their end model's predictive performance.

\section{Implementation Details}

\begin{figure}[h]
\centering
\shadowbox{\includegraphics[width=0.4\textwidth,clip]{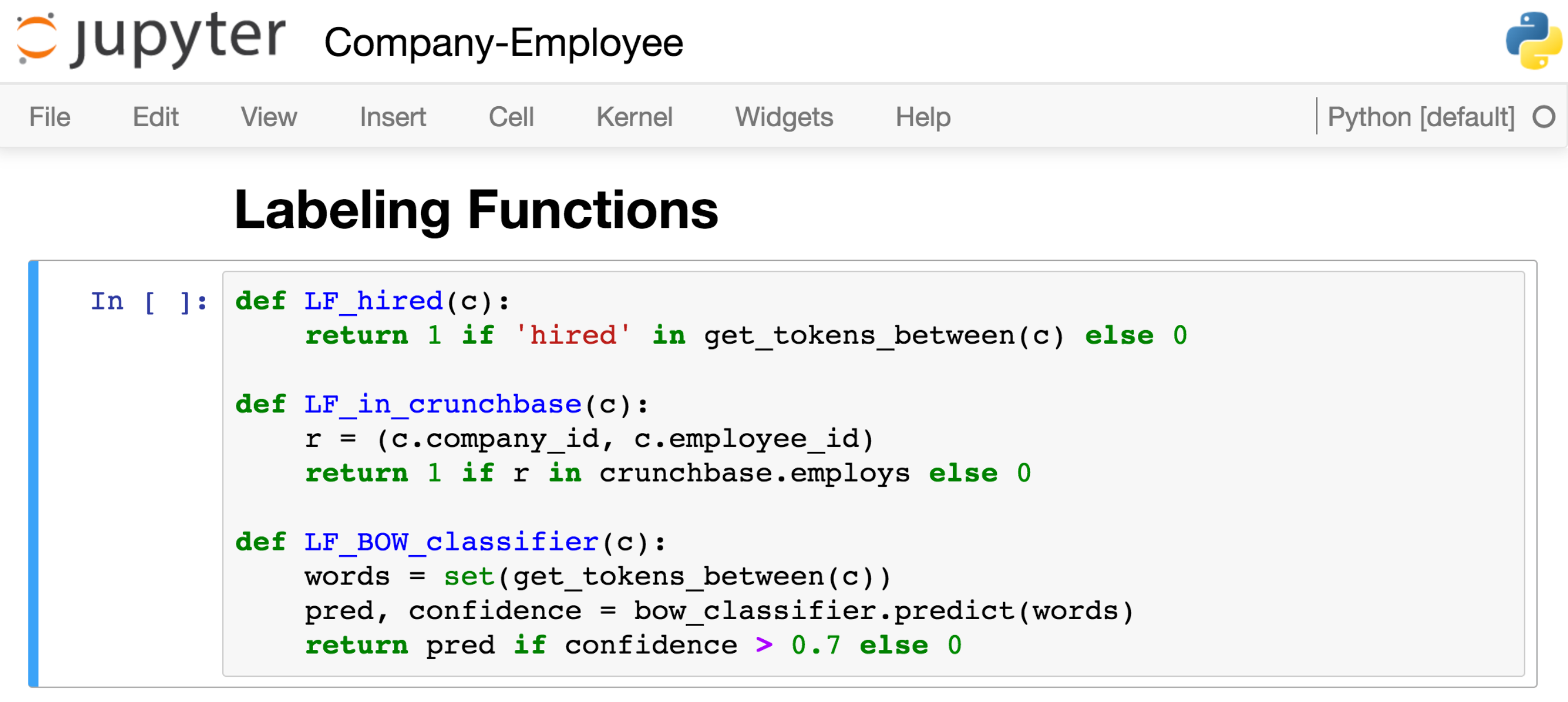}}
\caption{Labeling functions which express pattern-matching, distant supervision, and weak classifier heuristics, respectively, in Snorkel's Jupyter notebook interface.}
\label{fig:lfs}
\end{figure}

\begin{figure}[h]
\centering
\shadowbox{\includegraphics[width=0.35\textwidth,clip]{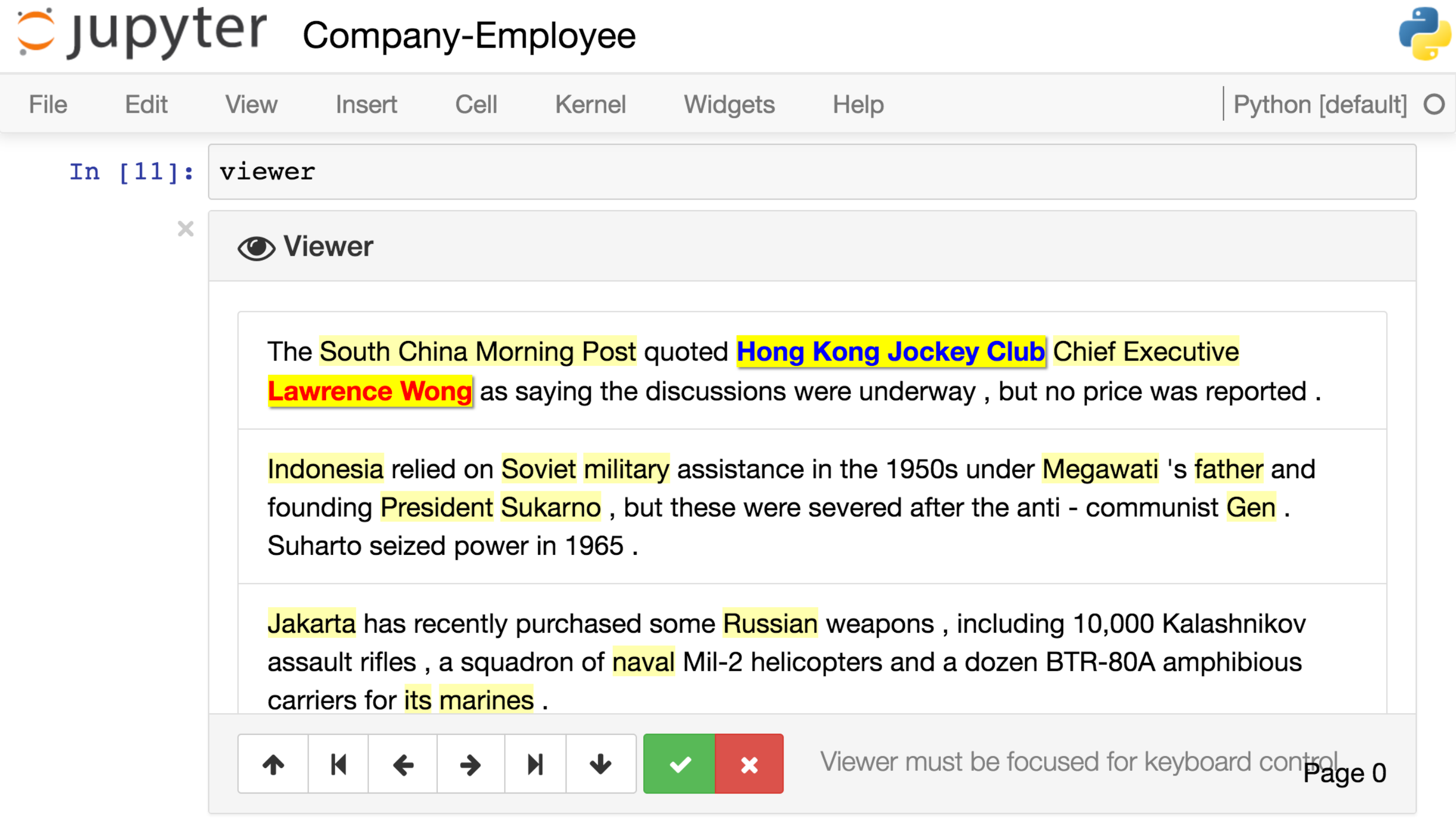}}
\caption{The Viewer utility in \systemx, showing candidate company-employee relation mentions from the ACE benchmark,
composed of candidate person and company mention pairs.}
\label{fig:viewer}
\end{figure}

Note that all code is open source and available---with tutorials, blog posts, workshop lectures, and other material---at \url{snorkel.stanford.edu}.

\paragraph*{Interface Implementation}

Snorkel's interface is designed to be accessible to SMEs without advanced programming skills.
All components run in Jupyter (\url{http://jupyter.org/}) iPython notebooks,
including writing labeling functions.
Users can therefore write labeling functions as arbitrary Python functions for maximum flexibility (Figure~\ref{fig:lfs}).
We also provide a library of labeling function primitives and generators to declaratively program weak supervision.

A key aspect of labeling function development is that the process is iterative.
After developing an initial set of labeling functions, it is important for users to visualize the errors of the end model.
Therefore, when the model is evaluated on the development data set, the candidates are separated into true positive, false positive, true negative, and false negative sets.
Each of these buckets can be loaded into a viewer in a notebook (Figure~\ref{fig:viewer}) so that SMEs can identify common patterns that are either not covered or misclassified by their current labeling functions.
The viewer also supports labeling candidates directly in order to create or expand development and test sets.

\paragraph*{Execution Model}

Since labeling functions are self-contained and operate on discrete candidates, their execution is embarrassingly parallel.
If Snorkel is connected to a relational database that supports simultaneous connections, e.g., PostgreSQL, then the master process (usually the notebook kernel) distributes the primary keys of the candidates to be labeled to Python worker processes.
The workers independently read from the database to materialize the candidates via the ORM layer, then execute the labeling functions over them.
The labels are returned to the master process which persists them via the ORM layer.
Collecting the labels at the master is more efficient than having workers write directly to the database, due to table-level locking.

Snorkel includes a Spark (\url{https://spark.apache.org/}) integration layer, enabling labeling functions to be run across a cluster.
Once the set of candidates is cached as a Spark data frame, only the closure of the labeling functions and the resulting labels need to be communicated to and from the workers.
This is particularly helpful in Snorkel's iterative workflow.
Distributing a large unstructured data set across a cluster is relatively expensive, but only has to be performed once.
Then, as users refine their labeling functions, they can be rerun efficiently.

This same execution model is supported for preprocessing \linebreak utilities---such as natural language processing for text and candidate extraction---via a common class interface.
Snorkel provides wrappers for Stanford CoreNLP (\url{https://stanfordnlp.github.io/CoreNLP/}) and SpaCy (\url{https://spacy.io/}) for text preprocessing, and supports automatically defining candidates using their named-entity recognition features.

\end{appendix}